\documentclass{article}

\usepackage[preprint, nonatbib]{neurips_2026}

\usepackage[utf8]{inputenc}      
\usepackage[T1]{fontenc}         
\usepackage{hyperref}            
\usepackage{url}                 
\usepackage{booktabs}            
\usepackage{amsfonts}            
\usepackage{amsmath}             
\usepackage{amssymb}             
\usepackage{nicefrac}            
\usepackage{microtype}           
\usepackage[table]{xcolor}       
\usepackage{graphicx}            
\usepackage{subcaption}          
\usepackage{multirow}            
\usepackage{makecell}            
\usepackage{array}               
\usepackage{longtable}           
\usepackage{rotating}            
\usepackage{enumitem}            
\usepackage{kotex}
\usepackage[
  backend=biber,
  style=ieee,
  sorting=nyt,
  doi=false,
  isbn=false,
  url=false,
  eprint=false,
]{biblatex}
\DeclareSourcemap{
  \maps[datatype=bibtex]{
    \map{
      \step[fieldsource=booktitle,
            match=\regexp{.*International\s+Conference\s+on\s+Learning\s+Representations.*},
            replace={International Conference on Learning Representations}]
    }
    \map{
      \step[fieldsource=booktitle,
            match=\regexp{.*AAAI\s+Conference\s+on\s+Artificial\s+Intelligence.*},
            replace={Proceedings of the AAAI Conference on Artificial Intelligence}]
    }
    \map{
      \step[fieldsource=booktitle,
            match=\regexp{.*Advances\s+in\s+Neural\s+Information\s+Processing\s+Systems.*},
            replace={Advances in Neural Information Processing Systems}]
    }
  }
}
\AtEveryBibitem{
  \clearfield{issn}
  \clearfield{note}
  \clearfield{series}
  \clearlist{publisher}
  \clearlist{location}
  \clearfield{address}
}
\newcommand{\dataset}{\textsc{SeoulMMOD}\xspace}
\usepackage{xspace}

\title{\dataset: A Large-Scale Multimodal Origin--Destination Flow Benchmark}

\author{%
  Taeyoung Yu\textsuperscript{1}\qquad
  Seonbin Jo\textsuperscript{2}\qquad
  Jiwon Kim\textsuperscript{1}\qquad
  Junyoung Byun\textsuperscript{3}\thanks{Corresponding author:
    \href{mailto:junyoungb@cau.ac.kr}{\texttt{junyoungb@cau.ac.kr}}.} \\
  \textsuperscript{1}The University of Queensland\qquad
  \textsuperscript{2}POSTECH\qquad
  \textsuperscript{3}Chung-Ang University
}

\begin{document}

\maketitle

\begin{abstract}
Origin-destination (OD) flow forecasting supports urban mobility decisions by predicting the volume and patterns of human movement across a city.
Most existing public OD benchmarks provide only a partial view of citywide mobility, covering a single transport service or at most two modes. 
With partial mode coverage, observed demand changes are difficult to separate into shifts across transport modes and changes in total travel. We introduce SeoulMMOD, a large-scale public benchmark for citywide multimodal mobility in Seoul.
SeoulMMOD provides three years of hourly OD flows estimating total mobility for 6 urban travel modes across 25 districts and 426 sub-districts, together with travel-time, travel-distance, calendar, rainfall, and point-of-interest (POI) signals. 
Using SeoulMMOD, we benchmark 14 forecasting baselines under a common protocol and analyse the effects of spatial scaling, joint versus mode-separate training, and cross-year generalization.
Results show that several spatio-temporal models that work at district level do not scale to sub-district OD forecasting, and that naive joint multi-mode training does not consistently improve accuracy. 
Together, the dataset and benchmark establish a reproducible testbed for multi-mode, multi-year OD forecasting at the city scale.
\end{abstract}


\section{Introduction}
\label{sec:intro}

Large cities move people across rail, bus, road, walking, and
shared-mobility systems every day. These movements affect congestion,
crowding, and service demand across the city. Urban transport
operators need forecasts of how many trips will occur from one zone to
another~\cite{rong_interdisciplinary_2024}.
An \textbf{origin-destination (OD) flow} counts trips from an origin
zone to a destination zone within a fixed time window. Collecting these
flows over all zone pairs gives an OD matrix, a standard representation
of citywide mobility.
\textbf{OD flow forecasting} predicts future trip volumes between zone
pairs from recent observations and feeds directly into operational
decisions such as
timetable optimization on congested rail
corridors~\cite{niu_optimizing_2013}, at-stop control
of bus networks to reduce bunching and overcrowding~\cite{gkiotsalitis_at-stop_2021},
and demand-aware rebalancing of shared-mobility fleets~\cite{sayarshad_non-myopic_2017}.

OD forecasting relies heavily on the availability of OD data.
OD flow is difficult to observe directly because trips are distributed across many transport modes and operators, and capturing every trip requires combining their separate records.
OD matrices have traditionally been built in two ways.
Household travel surveys~\cite{egu_how_2020} directly collect reported trips but have limited sample sizes and low update frequencies.
Roadside traffic counts~\cite{cascetta_estimation_1984} infer trip flows
from aggregated sensor data, but sparse coverage and non-unique
solutions limit these estimates.
Both produce static OD matrices and cannot capture how flows shift hour to hour or day to day.

Recent OD datasets are built from several operational data sources,
including smart-card and fare-gate logs from transit
agencies~\cite{hussain_transit_2021}, trip records from ride-hailing, taxi,
and bike-sharing platforms~\cite{wang_origin-destination_2019,yao_revisiting_2019},
GPS traces from taxi fleets~\cite{yuan2011t}, and
mobile-phone signalling data
from telecommunication carriers~\cite{iqbal_development_2014,wang_estimating_2013,alexander2015origin,fekih_data-driven_2021}. These
records include trip-level timestamps, which make it possible to
construct OD aggregates at hourly or finer resolution.

Many public OD datasets built from these sources provide only a
partial view of citywide mobility. Existing benchmarks usually cover a
single transport service or mode, and multi-mode datasets in the
forecasting literature cover at most two. With partial mode
coverage, observed demand changes are difficult to separate into shifts
across transport modes and changes in total travel. For instance, after
a rail service interruption, lower subway demand may reflect riders shifting to
buses or taxis, or it may reflect cancelled trips, each of which calls
for a different operational response. To our knowledge, no public OD
benchmark provides hourly citywide flows broken down by all major urban
transport modes.

\begin{figure}[t]
  \centering
  \includegraphics[width=\linewidth]{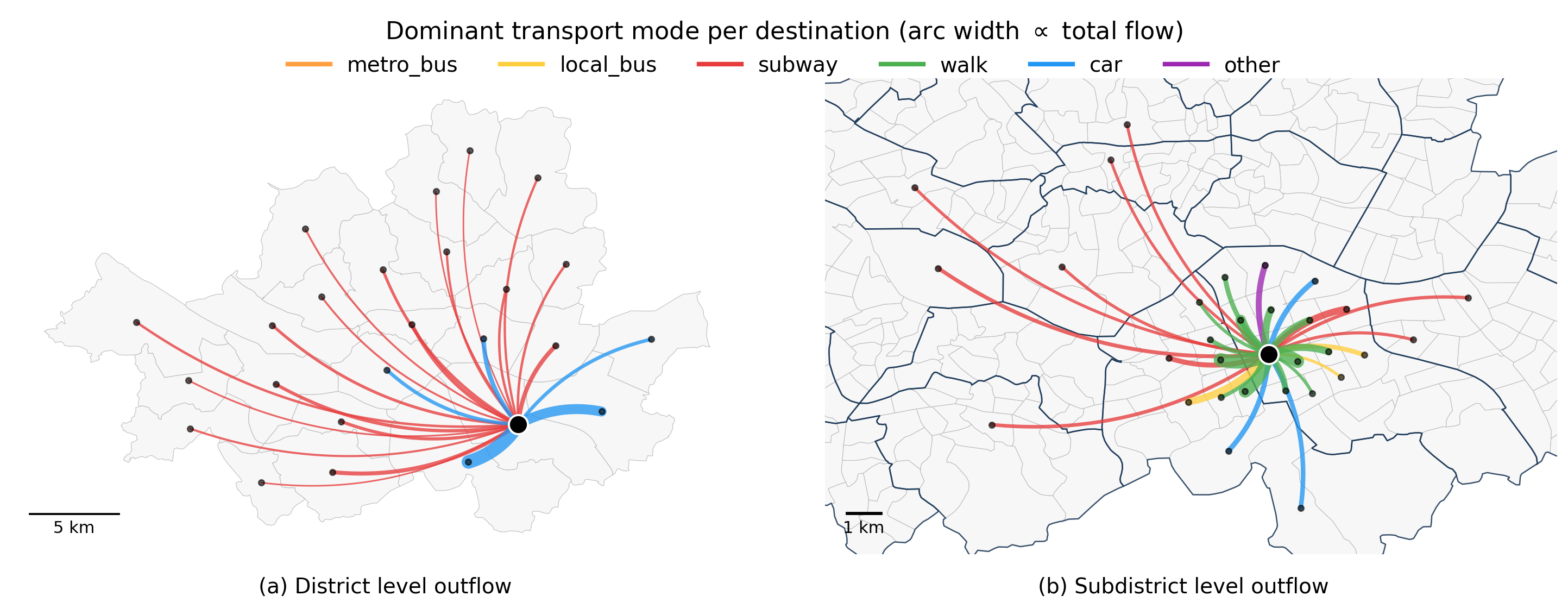}
  \vspace{-10pt}
  \caption{Annual OD outflow in \dataset (2024) from the same Seoul
    origin at district and sub-district levels. Arc color indicates the
    destination's dominant transport mode.}
  \label{fig:intro}
  \vspace{-8pt}
\end{figure}

To address these gaps, we introduce \dataset, an hourly citywide OD
benchmark covering all major urban transport modes.

\textbf{Contributions.}
\begin{enumerate}[leftmargin=*,itemsep=2pt,topsep=4pt]
  \item We release \dataset, the first public multi-mode time-series
  city-scale OD forecasting benchmark. It covers six transport
  modes over three years (2023--2025) at hourly resolution and two
  spatial levels (district and sub-district), and includes
  aligned travel-time, travel-distance, holiday, rainfall, and
  point-of-interest (POI) signals for broader mobility modeling.
  \item We benchmark 14 forecasting baselines on \dataset, providing
  reference performance and showing that several spatio-temporal
  models that work at district level do not scale to sub-district OD
  forecasting.
  \item We analyze joint training versus training one model per mode and
  cross-year generalization. Previous OD datasets do not support these
  evaluations under a unified multi-mode, multi-year protocol, as they
  typically cover either a single mode or a single year.
  The training comparison shows that naive multi-mode training
  does not consistently improve OD forecasting.
\end{enumerate}


\section{Related Work}
\label{sec:related}

\subsection{OD Forecasting}
\label{sec:related-models}

OD forecasting is a high-dimensional spatio-temporal forecasting
problem, where each time step is represented by a $Z \times Z$ OD
matrix. The future flow of each OD pair shares temporal patterns with
related pairs, such as nearby origins or competing destinations. Deep
learning models have been used to capture temporal dynamics
and dependencies among OD pairs. Toqué et
al.~\cite{toque_forecasting_2016} and Liu et
al.~\cite{liu_contextualized_2019} use LSTMs for public-transport and
taxi OD forecasting. To model spatial interactions in OD matrices,
Wang et al.~\cite{wang_origin-destination_2019} use grid embeddings,
Jiang et al.~\cite{jiang_countrywide_2021} use two-dimensional graph
convolutions, and Huang et al.~\cite{huang_odformer_2023} use attention
mechanisms across OD pairs. Liu et al.~\cite{liu_odmixer_2025} use an MLP-based architecture
with separate modules for temporal mixing, origin/destination-view
spatial mixing, and trend modeling.

Recent work predicts mobility demand across multiple transport modes
jointly. Zhang et al.~\cite{zhang_joint_2026} jointly predict subway-bus
and subway-bike OD demand and report improvements over single-mode
variants. At the zone level rather than the OD-pair level, Liang et
al.~\cite{liang_joint_2022} jointly predict subway and ride-hailing
inflow/outflow demand with multi-relational GNNs, and Ke et
al.~\cite{ke_joint_2021} jointly predict solo and shared ride-hailing
demand with multi-task multi-graph learning. Together, these studies
show that cross-mode signals can improve forecasting, but existing
studies typically use data that span less than one year and cover at
most two transport modes.

\subsection{Existing OD Datasets}
\label{sec:related-datasets}

Public OD datasets fall into two categories. Static OD datasets
provide a single time-invariant OD matrix per zone pair, often
representing commuting flows between residential and workplace regions.
For example, LEHD Origin-Destination Employment Statistics (LODES)
dataset\footnote{\url{https://lehd.ces.census.gov/data/}} provides annual
home-to-work commuting counts that have been used to build commuting
OD datasets at U.S.\ county and New York City census-tract
scales~\cite{rong_large-scale_2025,liu_learning_2020}.
A single annual snapshot cannot represent hourly demand swings,
weekday-weekend differences, or seasonal effects needed for
operational forecasting.

Dynamic OD datasets provide OD matrices that change over time,
often aggregated from trip records, smart-card taps, or other
time-stamped mobility records at hourly or finer resolution. Forecasting
studies have used ride-hailing OD matrices from ride-hailing platforms
such as UCAR and Didi~\cite{wang_origin-destination_2019}, smart-card
metro OD in Chinese cities~\cite{liu_online_2023}, and subway-bus or
subway-bike OD data~\cite{zhang_joint_2026}. Ma et
al.~\cite{ma_forecasting_2022}
aggregate public NYC taxi and bike trip records into regional inflow,
outflow, and OD matrices.

Existing dynamic OD datasets share several limitations. First, public
datasets typically reflect a single operator, and multi-mode forecasting
studies cover at most two modes. In some studies that combine records
across operators, modes are not aggregated to a common spatial unit,
requiring additional alignment.
Second, existing public benchmarks remain small relative to forecasting
all OD pairs across a city. Success on a station network or
coarse grid does not imply that a method scales to finer citywide
administrative zones.
Third, temporal spans of weeks to
months make it hard to evaluate seasonal patterns or cross-year transfer.
Table~\ref{tab:datasets-od} summarises these datasets alongside
\dataset.

\begin{table}[t]
  \centering
  \small
  \renewcommand{\arraystretch}{1.15}
  \caption{Related dataset comparison. \emph{Trip attrs} lists
    per-record auxiliary attributes (average travel time / distance).
    \emph{Context} lists bundled signals (holiday, point-of-interest, weather).
    ``+'' denotes separate node sets for different modes.}
  \label{tab:datasets-od}
  \setlength{\tabcolsep}{3pt}
  \resizebox{\textwidth}{!}{%
  \begin{tabular}{ll l ll l l c}
    \toprule
    Dataset & Coverage & Modes & \# Nodes & \# OD pairs & Trip attrs & Context & Public \\
    \midrule
    \multicolumn{8}{l}{\emph{Zone-level forecasting}} \\
    \midrule
    Liang et al.~\cite{liang_joint_2022}          & 2018-03 -- 2018-08 & subway, ride-hailing                          & 136 + 63      & ---       & $\times$ & $\times$ & $\times$   \\
    Ke et al.~\cite{ke_joint_2021}           & 2018-01 -- 2018-12 & \makecell[l]{solo ride-hailing,\\shared ride-hailing} & 63            & ---       & $\times$ & $\times$ & $\times$   \\
    Yao et al.~\cite{yao_revisiting_2019}                  & 2015-01 -- 2015-03 / 2016-07 -- 2016-08 & bike, taxi                                     & 200           & ---       & $\times$ & $\times$ & \checkmark \\
    \midrule
    \multicolumn{8}{l}{\emph{OD-level forecasting}} \\
    \midrule
    Zhang et al.~\cite{zhang_joint_2026} (BJ)   & 2016-02 -- 2016-04 & subway, bus                                   & 276 + 2{,}320 & 28{,}802  & $\times$ & $\times$ & $\times$   \\
    Zhang et al.~\cite{zhang_joint_2026} (NYC)  & 2024 (84d, sampled) & subway, bike                                  & 121 + 1{,}081 & 48{,}474  & $\times$ & $\times$ & $\times$   \\
    UCAR~\cite{wang_origin-destination_2019}                    & 2016-08 (31d)      & ride-hailing                                  & 400           & 160{,}000 & $\times$ & \makecell[l]{Wthr.,\\POI} & $\times$   \\
    DIDI~\cite{wang_origin-destination_2019}                    & 2016-11 (30d)      & ride-hailing                                  & 225           & 50{,}625  & $\times$ & $\times$ & $\times$   \\
    Deep-NYC-Taxi-Bike~\cite{ma_forecasting_2022} & 2019-01 -- 2020-12 & taxi, bike                                    & 69            & 4{,}761   & $\times$ & $\times$ & \checkmark \\
    HZMOD~\cite{liu_online_2023}                    & 2019-01 (25d)      & subway                                        & 80            & 6{,}400   & $\times$ & $\times$ & \checkmark \\
    SHMOD~\cite{liu_online_2023}                    & 2016-07 -- 2016-09 & subway                                        & 288           & 82{,}944  & $\times$ & $\times$ & \checkmark \\
    \midrule
    \rowcolor{gray!12}
    \textbf{\dataset (ours)} & \textbf{2023-01 -- 2025-12} & \textbf{\makecell[l]{metro bus, local bus, subway,\\walk, car, other}} & \textbf{25 / 426} & \textbf{625 / 181{,}476} & \textbf{\makecell[l]{Avg time,\\Avg dist}} & \textbf{\makecell[l]{Hol., POI,\\Wthr.}} & \checkmark \\
    \bottomrule
  \end{tabular}%
  }
\end{table}


\section{\dataset Dataset}
\label{sec:dataset}

\subsection{Data Source and Collection}
\label{sec:dataset-source}

We build \dataset from publicly available ``Seoul Metropolitan
Area Living Mobility Data,'' jointly produced by the Seoul Metropolitan
Government and KT (a major South Korean
mobile carrier) and distributed through Seoul's open-data
portal\footnote{\url{https://data.seoul.go.kr}}.
According to the public source documentation, the records are derived from KT
LTE/5G cell-tower signalling logs, a well-established source for human mobility
analysis~\cite{gonzalez_understanding_2008} and OD-demand estimation
validated against travel surveys~\cite{toole_path_2015}. The original source
aggregates these records into administrative-zone OD counts by
mode and time, with population-scaling corrections that aim to
reduce estimation error from relying on one carrier's subscriber base. Each record reports
an estimated total trip volume together with the average travel
distance and average travel time for that OD pair, mode, and time
bin, auxiliary attributes rarely included in previous public OD
benchmarks. Appendix~\ref{app:datasheet} summarises the original source
generation procedure described in the public source documentation and documents our
preprocessing steps.

To construct \dataset, we filter the original source records to OD pairs
whose origin and destination both lie within Seoul. We keep the six
intra-urban modes (metro bus, local bus, subway, walk, car, other) and
drop the three intercity modes (air, train, express bus). We reconcile
administrative-code changes during 2023--2025 so every
sub-district has a stable identity, and unify the source time bins to
a uniform 1\,h grid.
After these steps, \dataset spans 2023-01-01 through 2025-12-31,
covering 1{,}096 days and 26{,}304 hourly time-steps at two spatial
levels, 25 districts and 426 sub-districts within Seoul.\footnote{The original public data are distributed under the
Korea Open Government License (KOGL) Type~1, which permits
unrestricted use, redistribution, modification, and commercial use
subject only to source attribution. See
\url{https://www.kogl.or.kr/info/licenseTypeEn.do}.}

\subsection{Dataset Analysis}
\label{sec:dataset-analysis}

\dataset shows uneven spatial demand, recurrent temporal cycles, and
mode-specific network structure.

\begin{figure}[t]
  \centering
  \begin{subfigure}[t]{0.24\textwidth}
    \includegraphics[width=\textwidth]{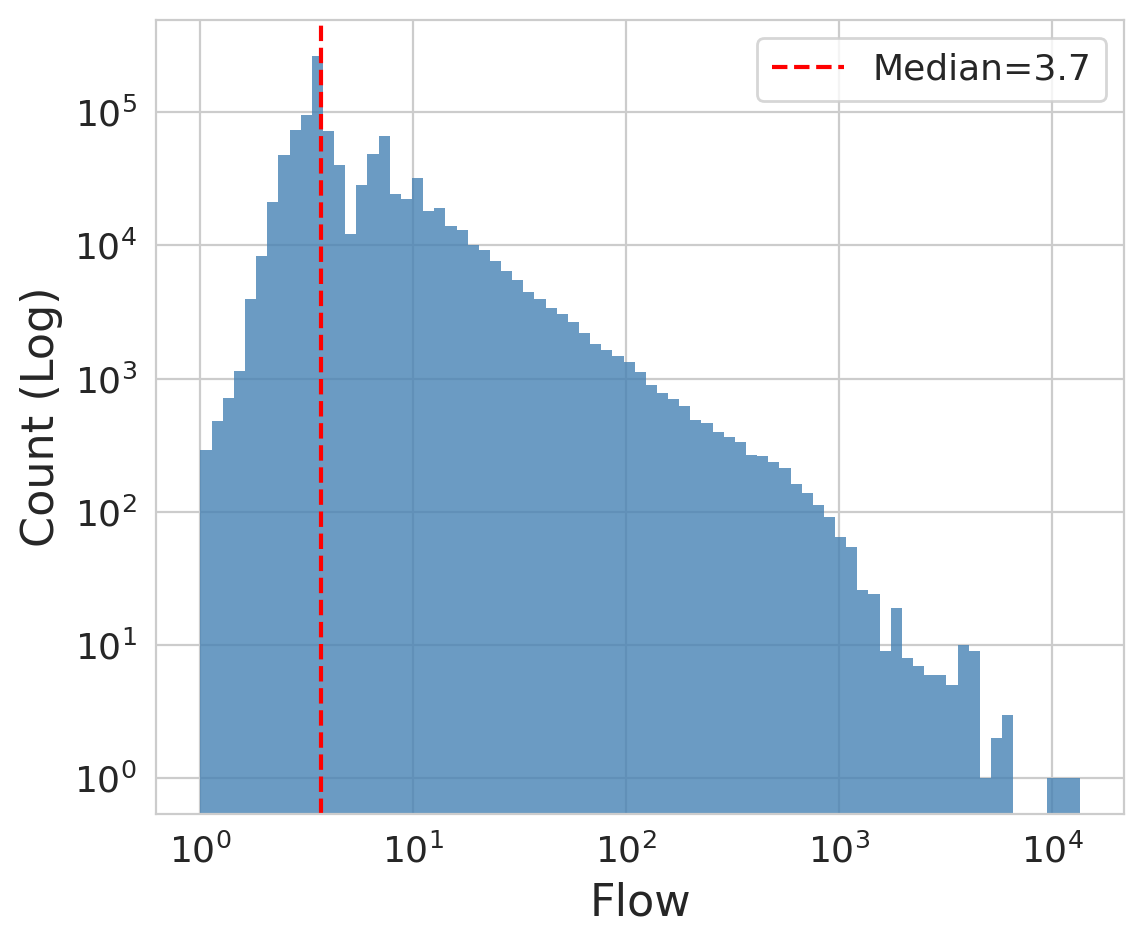}
    \caption{Flow distribution}
    \label{fig:flow-density}
  \end{subfigure}\hfill
  \begin{subfigure}[t]{0.24\textwidth}
    \includegraphics[width=\textwidth]{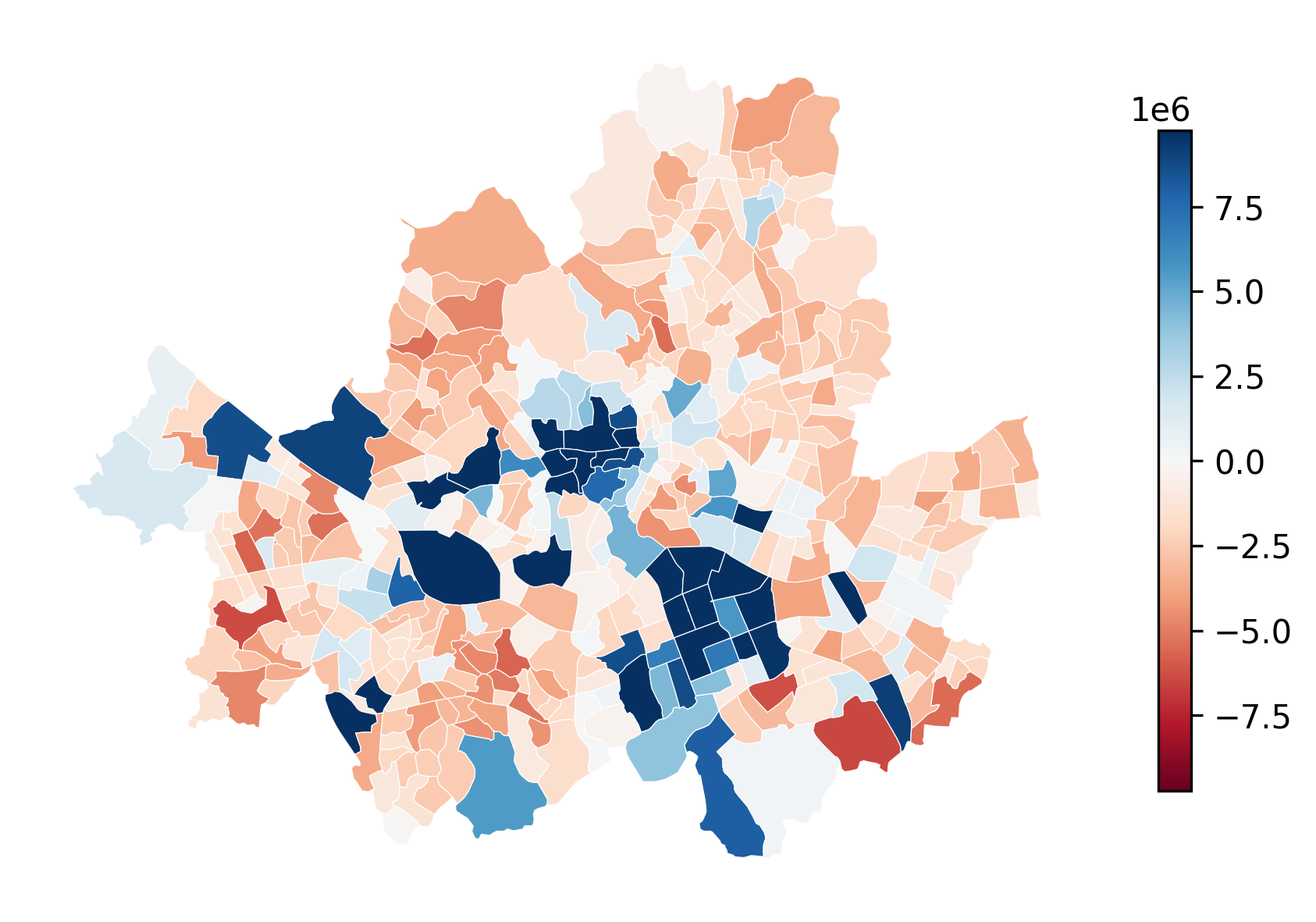}
    \caption{Net inflow}
    \label{fig:net-inflow}
  \end{subfigure}
  \hfill
  \begin{subfigure}[t]{0.24\textwidth}
    \includegraphics[width=\textwidth]{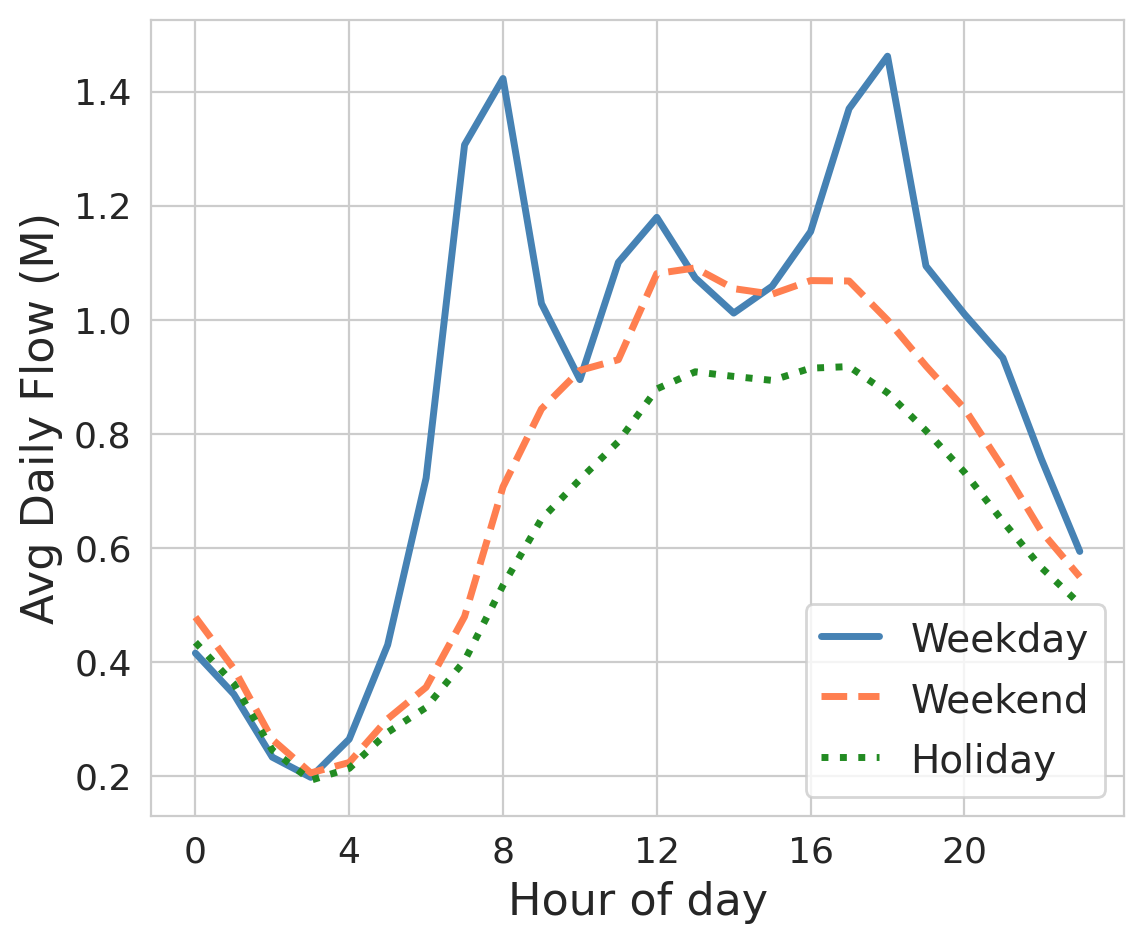}
    \caption{24-hour profile}
    \label{fig:24h}
  \end{subfigure}\hfill
  \begin{subfigure}[t]{0.24\textwidth}
    \includegraphics[width=\textwidth]{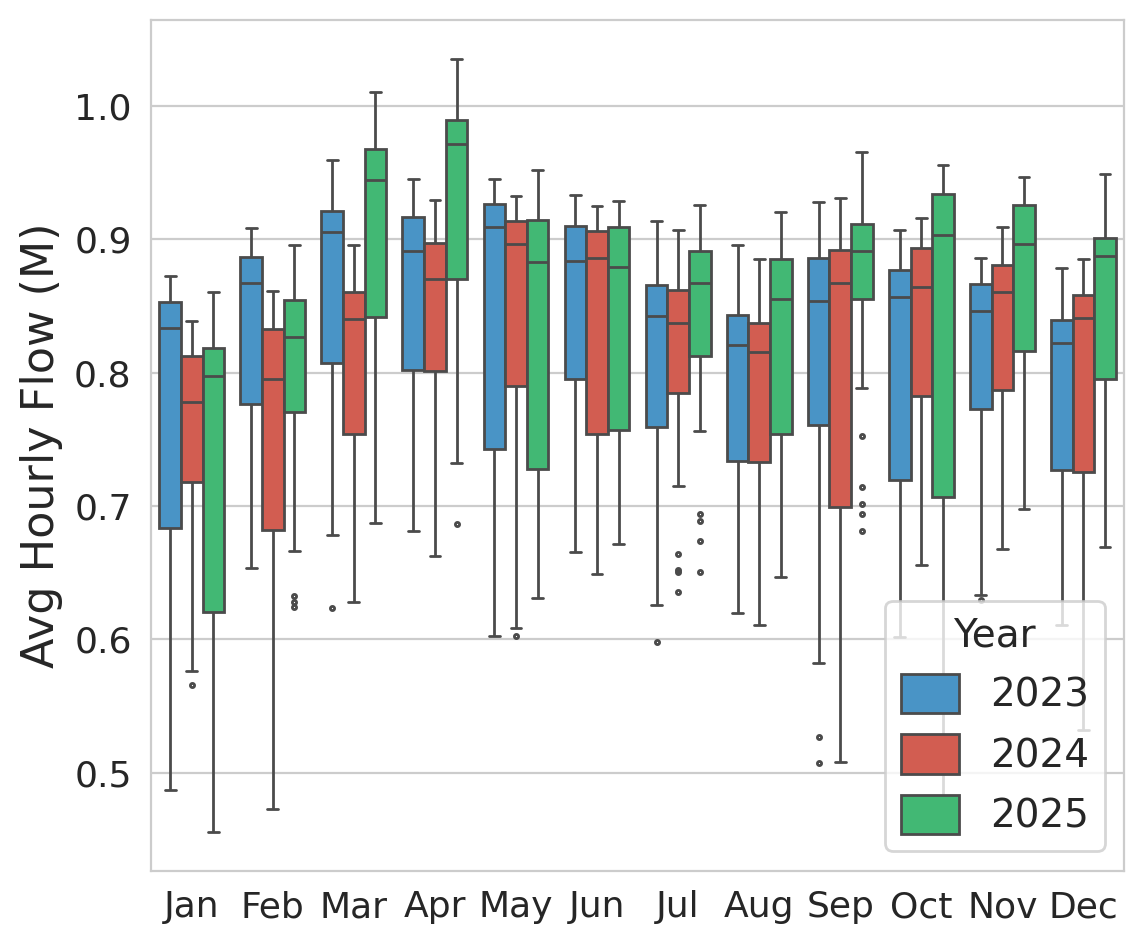}
    \caption{Monthly trend}
    \label{fig:monthly}
  \end{subfigure}

  \caption{Aggregate and temporal characteristics of \dataset.
    (a)~Distribution of OD-record flow.
    (b)~Morning-rush net inflow, computed as inflow minus outflow
    during 07:00--09:00.
    (c)~Hourly profile by day type.
    (d)~Monthly trend across years.}
  \label{fig:overview-temporal}
\end{figure}

\textbf{Spatio-temporal dynamics.}
\dataset varies across space, time, and transport modes.
OD flow follows a heavy-tailed distribution
(Figure~\ref{fig:flow-density}), consistent with observed human
mobility patterns in cities~\cite{gonzalez_understanding_2008}.
The spatial pattern is uneven. During morning rush hour,
net inflow separates destination-heavy sub-districts from
origin-heavy sub-districts (Figure~\ref{fig:net-inflow}).
Figure~\ref{fig:24h} shows how flow changes over the day. Peaks occur
around 8:00 and 18:00 due to commuting, and disappear on holidays.
Total flows vary by month and year (Figure~\ref{fig:monthly}),
showing longer-term shifts beyond the daily commuting cycle.


\begin{figure}[t]
  \centering
  \begin{subfigure}[t]{0.24\textwidth}
    \includegraphics[width=\textwidth]{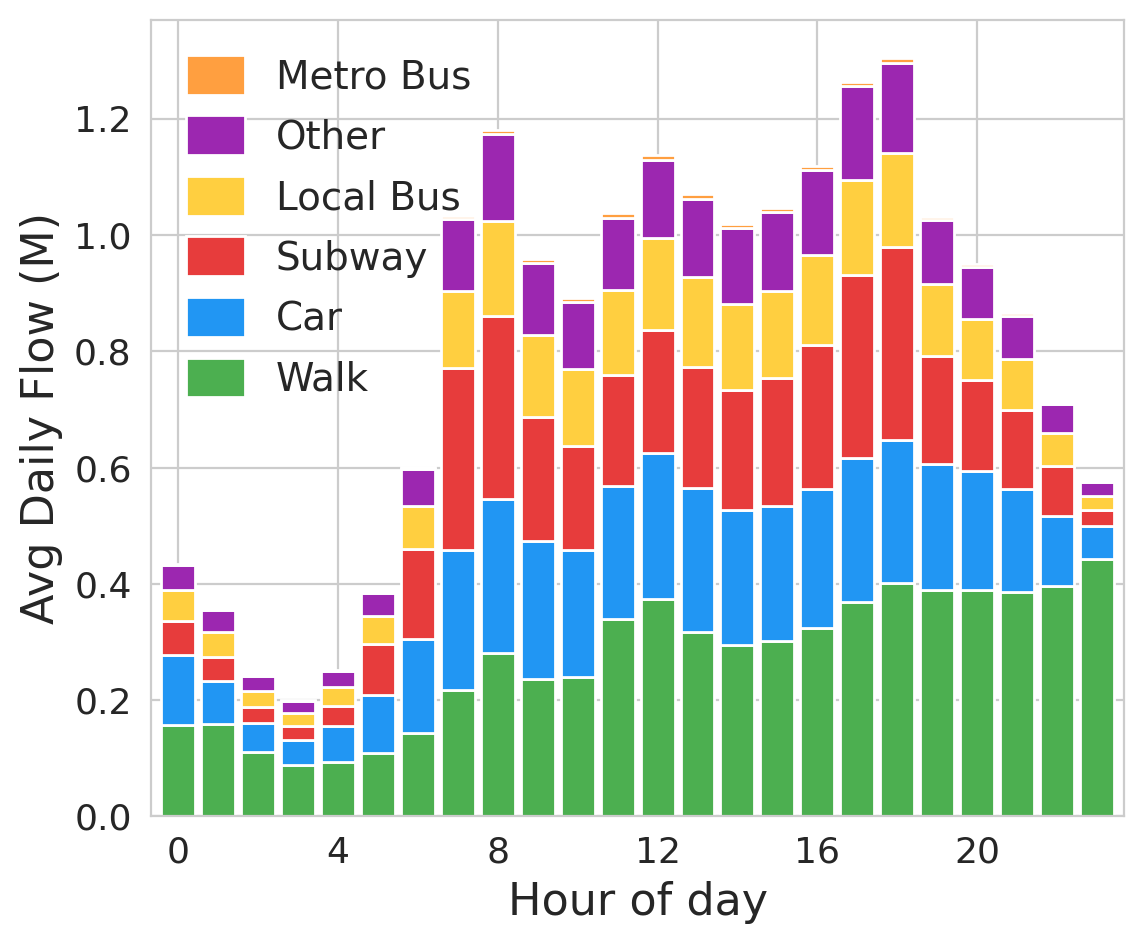}
    \caption{Mode composition}
    \label{fig:mode-comp}
  \end{subfigure}\hfill
  \begin{subfigure}[t]{0.24\textwidth}
    \includegraphics[width=\textwidth]{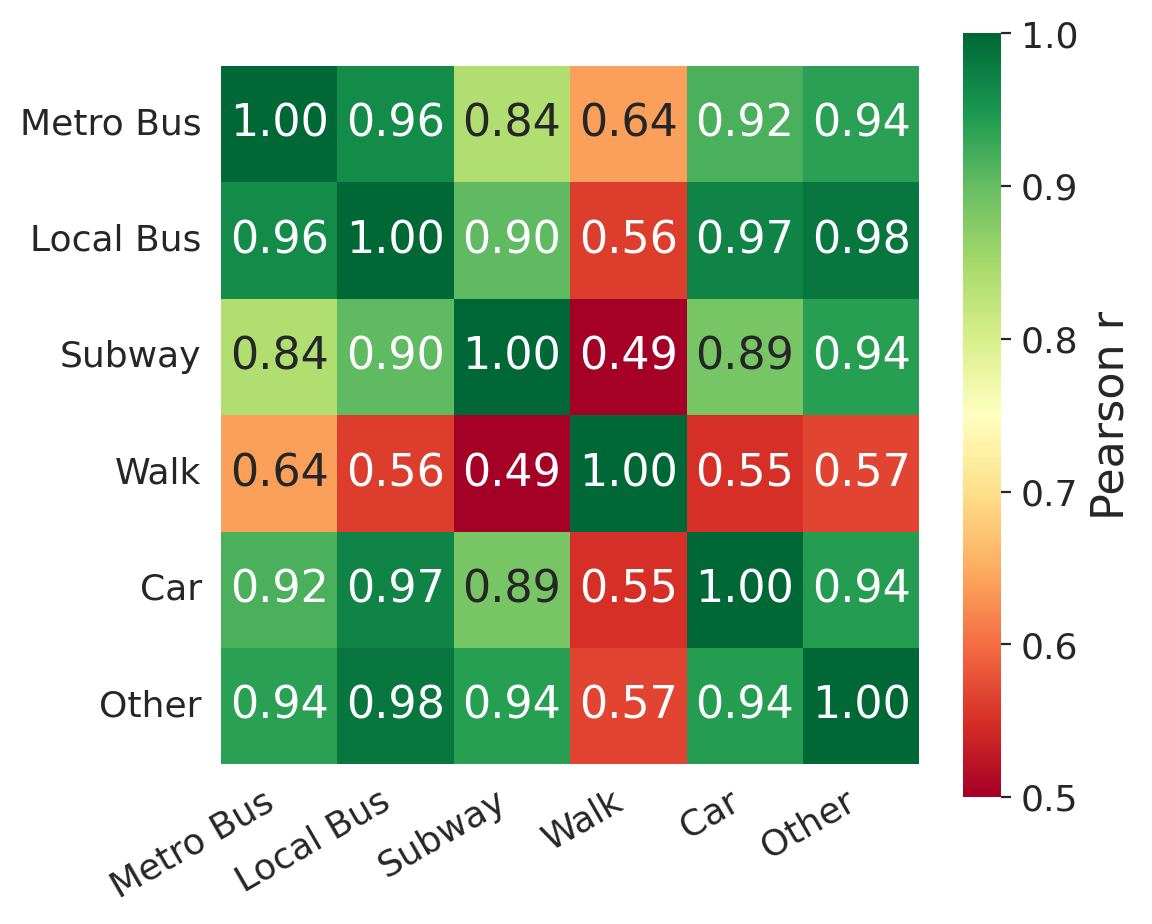}
    \caption{Temporal correlation}
    \label{fig:temp-corr}
  \end{subfigure}\hfill
  \begin{subfigure}[t]{0.24\textwidth}
    \includegraphics[width=\textwidth]{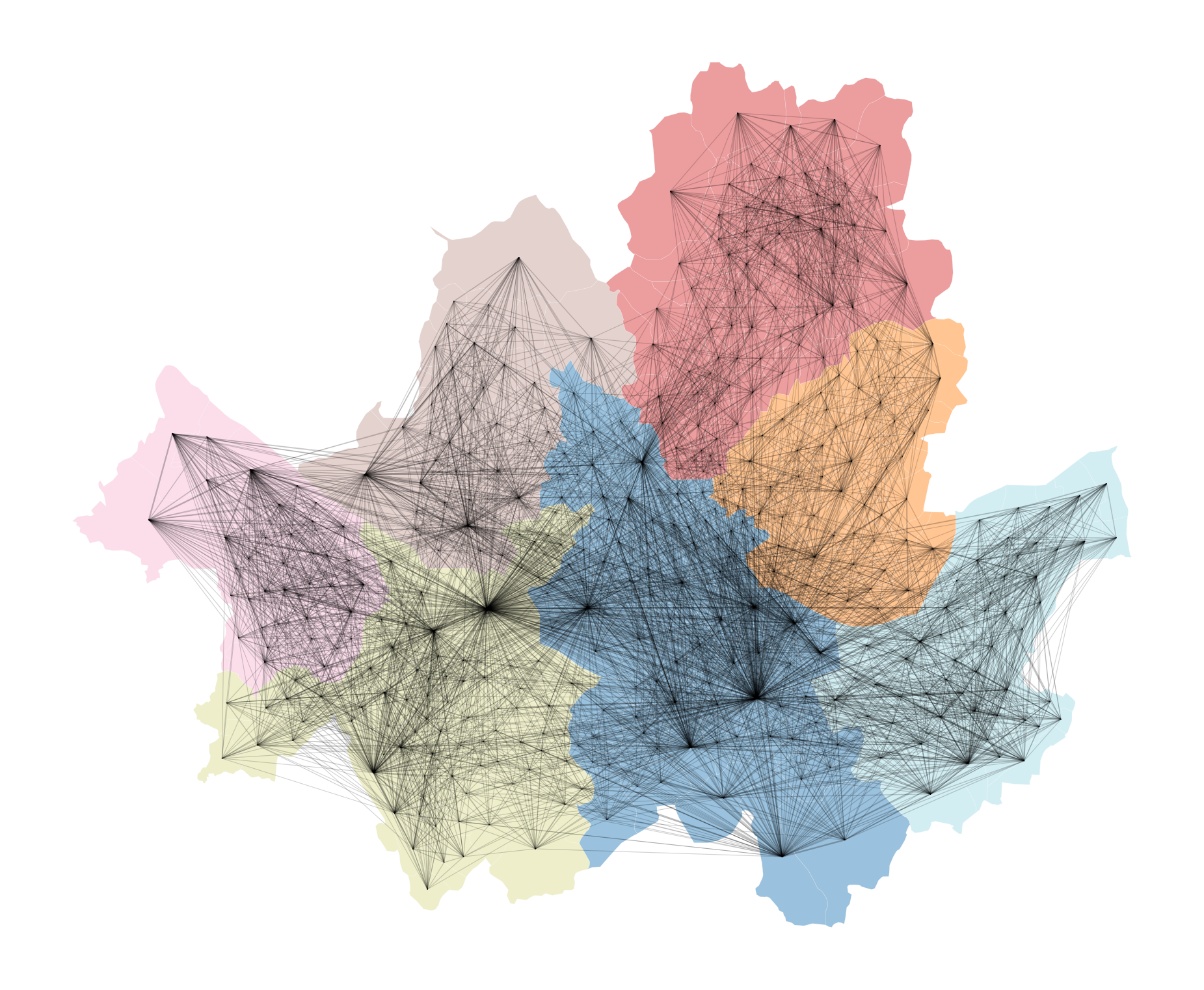}
    \caption{Car backbone}
    \label{fig:net-car}
  \end{subfigure}\hfill
  \begin{subfigure}[t]{0.24\textwidth}
    \includegraphics[width=\textwidth]{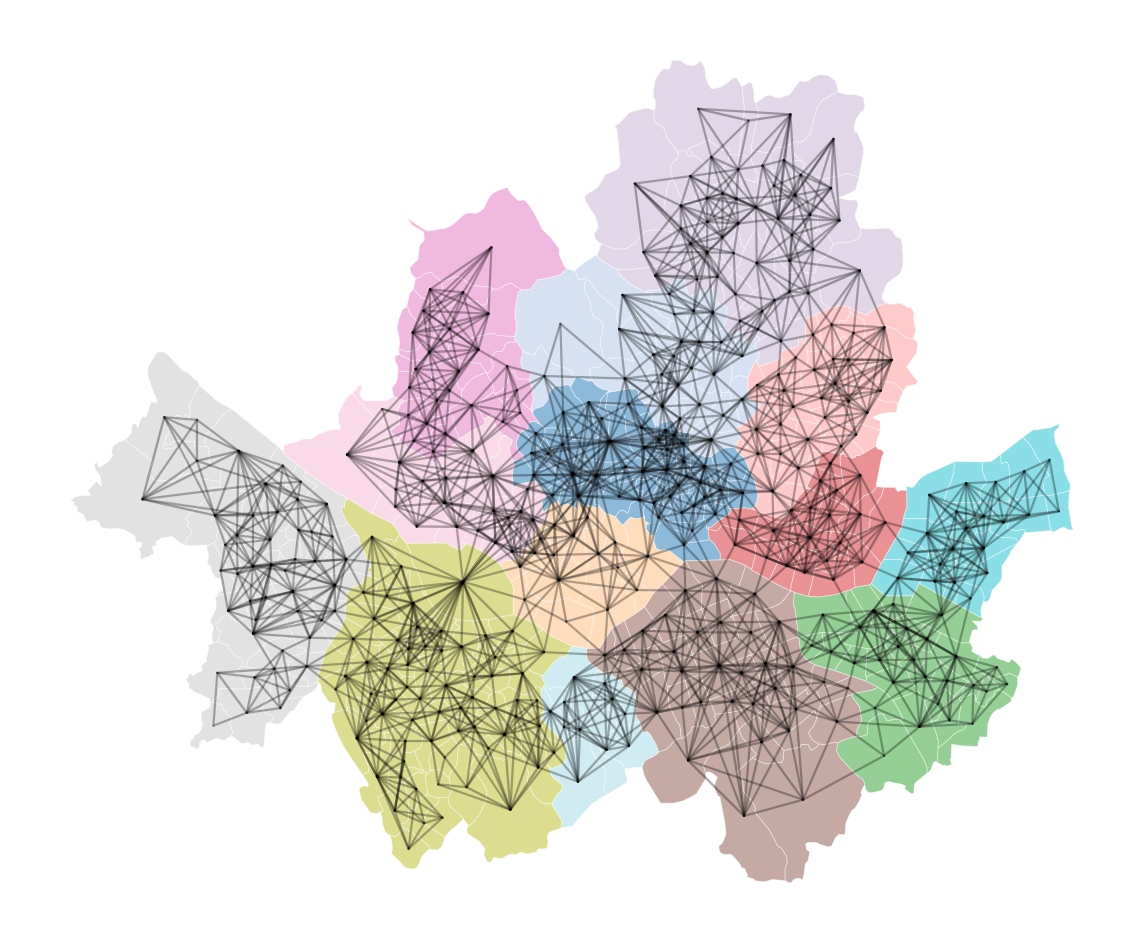}
    \caption{Walk backbone}
    \label{fig:net-walk}
  \end{subfigure}

  \caption{Multi-mode structural diversity.
    (a)~Mode composition over the day.
    (b)~Pairwise Pearson correlation of hourly total flow between modes.
    (c,\,d)~Disparity-filtered backbone networks and Louvain communities
    for car and walk, with edge opacity and width proportional to flow.
    Additional distributional and community-similarity views are shown
    in Appendix~\ref{app:additional-dataset-analysis}.}
  \label{fig:multimode}
\end{figure}

\textbf{Multi-mode structural diversity.}
The six modes show different daily activity profiles
(Figure~\ref{fig:mode-comp}) and different temporal correlations
(Figure~\ref{fig:temp-corr}). For instance, Walk becomes dominant late
at night, when other transport modes are unavailable. Additional
distributional views in Appendix~\ref{app:additional-dataset-analysis}
show that all modes are heavy-tailed, but differ in OD-record flow
scale and distance.
To examine whether these differences extend to spatial organisation,
we construct a network from each mode's OD matrix, apply the
disparity filter to extract the backbone~\cite{serrano_extracting_2009},
and run Louvain community detection~\cite{blondel_fast_2008}. Figures~\ref{fig:net-car}
and~\ref{fig:net-walk} show the results.
Car trips show long-range connections spanning the city, which leads to
low modularity and few large clusters ($Q{=}0.584$, 7~clusters).
In contrast, walk trips show short-range connections that form many
localised clusters ($Q{=}0.852$, 14~clusters).
The modes differ in spatial community structure, indicating that each
mode connects Seoul's zones in a different way.

\subsection{Auxiliary Data}
\label{sec:dataset-context}

\begin{figure}[t]
  \centering
  \begin{subfigure}[t]{0.24\textwidth}
    \includegraphics[width=\textwidth]{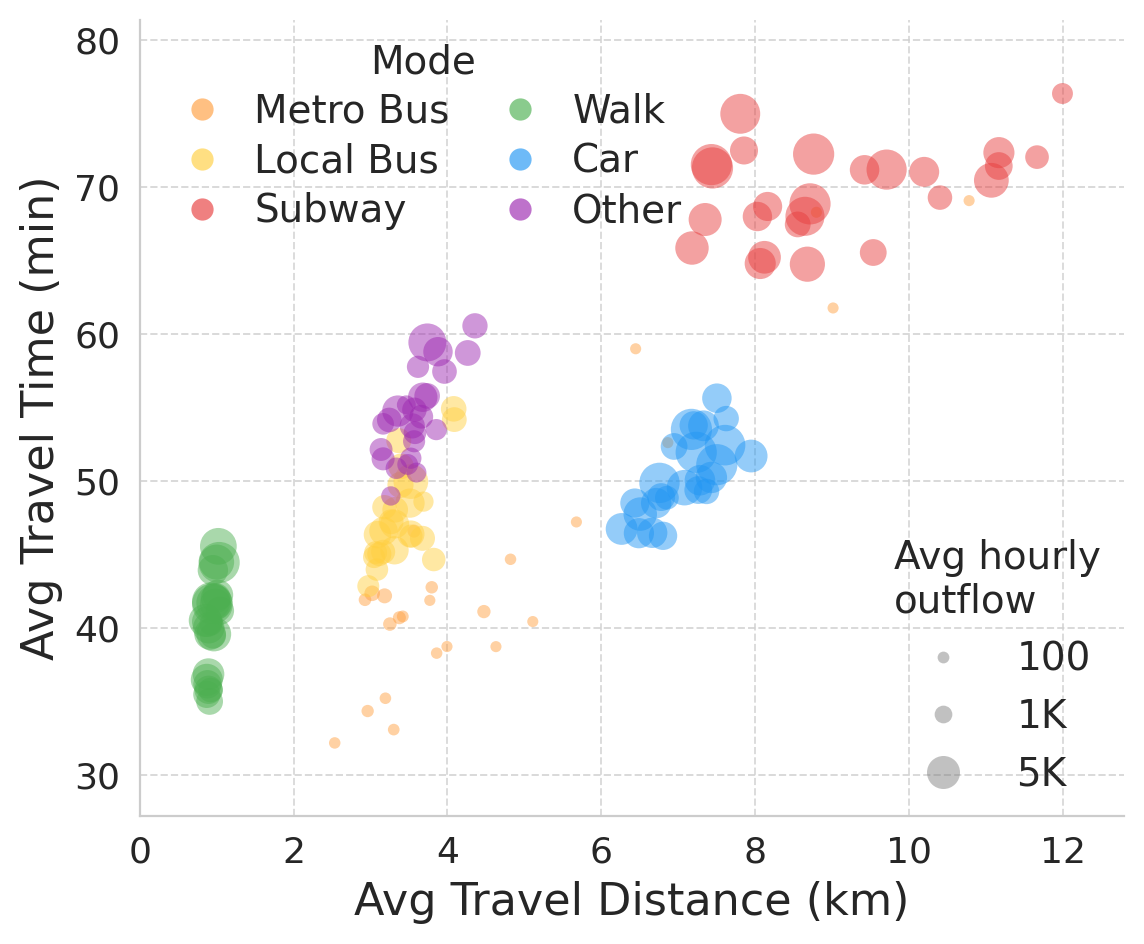}
    \caption{Mode attributes}
    \label{fig:context-time-outflow}
  \end{subfigure}\hfill
  \begin{subfigure}[t]{0.24\textwidth}
    \includegraphics[width=\textwidth]{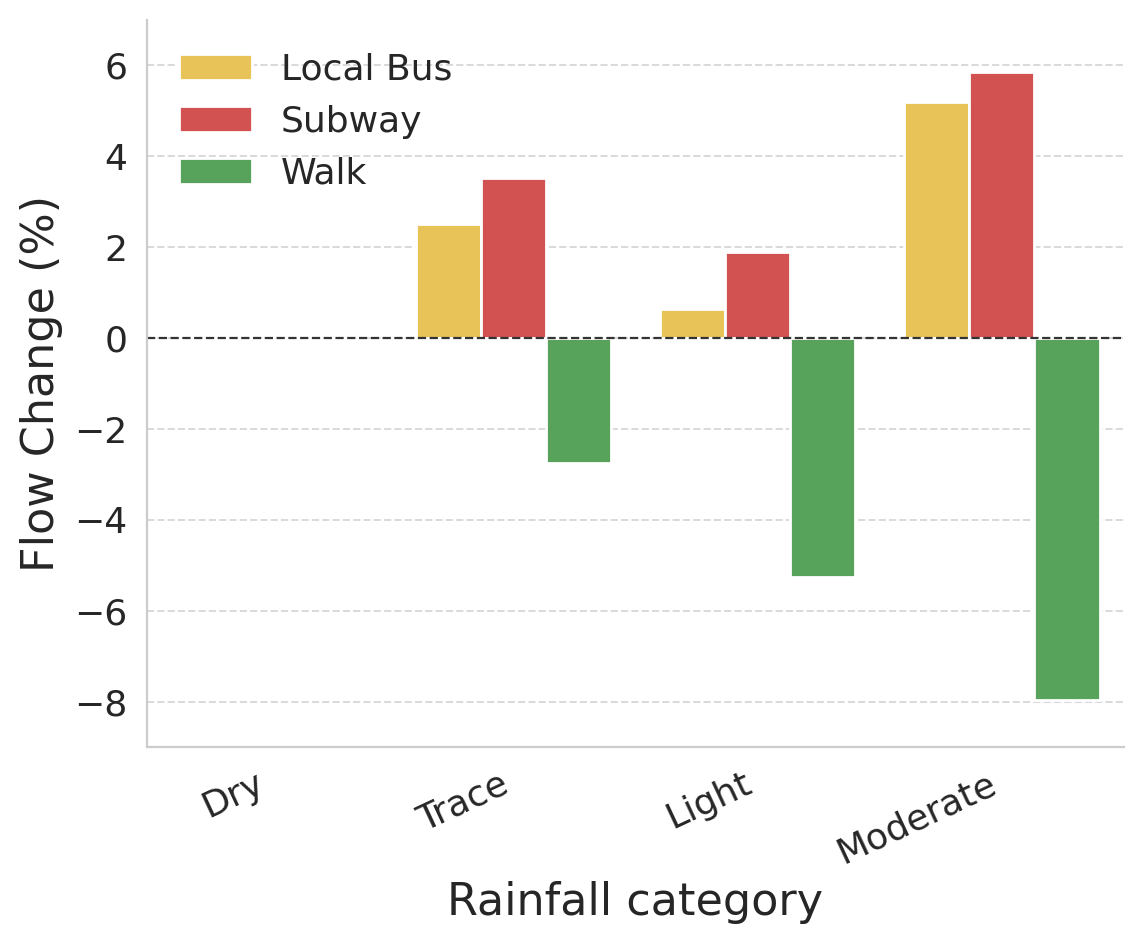}
    \caption{Rainfall by mode}
    \label{fig:context-rain}
  \end{subfigure}\hfill
  \begin{subfigure}[t]{0.24\textwidth}
    \includegraphics[width=\textwidth]{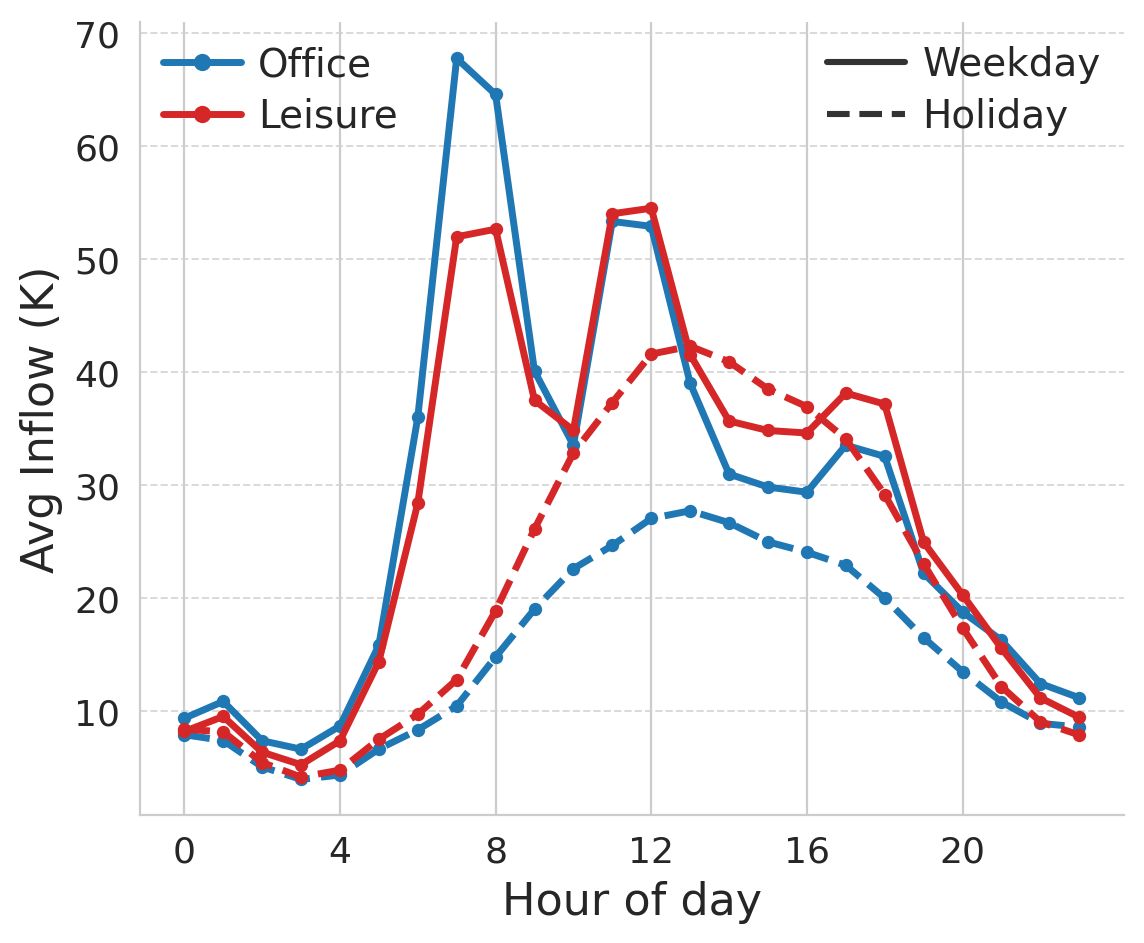}
    \caption{Calendar shift}
    \label{fig:context-calendar}
  \end{subfigure}\hfill
  \begin{subfigure}[t]{0.24\textwidth}
    \includegraphics[width=\textwidth]{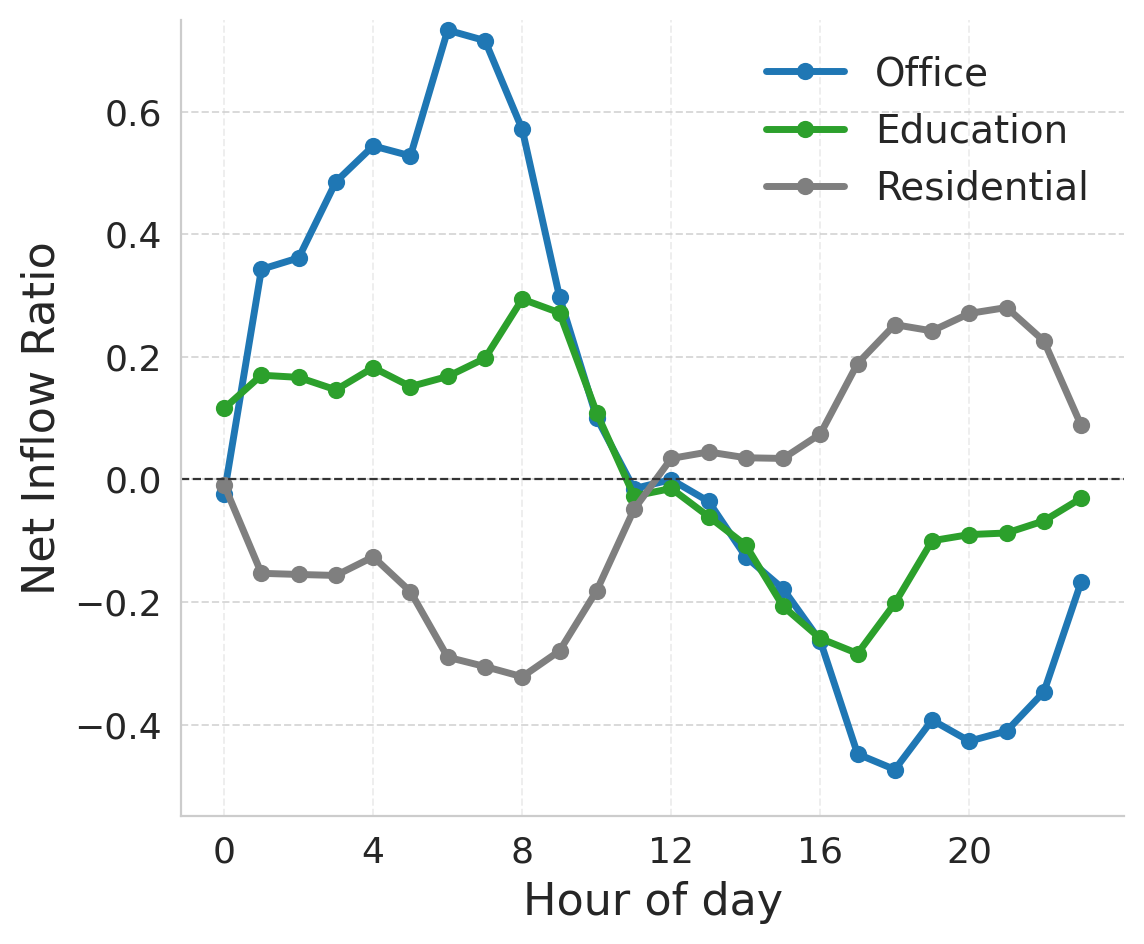}
    \caption{POI examples}
    \label{fig:context-poi}
  \end{subfigure}

  \caption{Trip attributes and contextual information in \dataset.
    (a)~District-level travel distance and travel time, with marker size
    showing average hourly outflow.
    (b)~Per-mode hourly flow change under rainfall categories,
    relative to dry hours.
    (c)~Hourly inflow for office and leisure destination examples on
    weekdays and holidays.
    (d)~Weekday net-inflow ratio for selected office, education, and
    low-POI residential examples.}
  \label{fig:context-auxiliary}
\end{figure}

\dataset provides auxiliary data rarely included in previous public OD
benchmarks, including trip attributes and contextual signals.
Figure~\ref{fig:context-auxiliary}
shows representative descriptive views of these fields, and
Appendix~\ref{app:context-analysis} provides further descriptive and
predictive analyses.

\textbf{Trip attributes.} Each OD record includes average travel time
and average travel distance for the corresponding OD pair, mode, and
hour. Figure~\ref{fig:context-auxiliary}(a) relates these attributes to
average hourly outflow, showing that each mode concentrates its larger
flows in different travel-distance and travel-time ranges.

\textbf{Calendar.} A daily table aligned with the OD time index records
day of week, weekend/weekday flag, Korean public-holiday flag, and
holiday name. The table covers all 1{,}096 days of the dataset.
Figure~\ref{fig:context-auxiliary}(c) shows how weekday and holiday
inflow profiles differ. Office destinations have a sharp weekday
morning peak that is much weaker on holidays, while leisure
destinations shift toward broader daytime activity.

\textbf{Rainfall.} Hourly precipitation totals at district level
for the full 2023--2025 window, sourced from the Seoul Metropolitan
rainfall observation network and distributed through the Seoul Open
Data Plaza. Figure~\ref{fig:context-auxiliary}(b) shows per-mode flow
changes under rainfall categories, with lower walk flow and higher
local-bus and subway flow on rainy hours.

\textbf{POI categories.} Per-sub-district counts of points of interest
from the Seoul Open Data Plaza commercial-district POI release. We use
the 2025 snapshot, and the source is updated once per year. These
counts cover 20 categories, spanning transport infrastructure,
commerce and leisure, public and health services, and education.
Figure~\ref{fig:context-auxiliary}(d) compares weekday net-inflow roles
for selected office, education, and low-POI residential examples.
Office and education examples receive net inflows in the morning and
shift toward net outflows later in the day, while the residential
example follows the reverse pattern.

\section{Experiments}
\label{sec:experiments}

We use \dataset to investigate how existing forecasting models behave
across multiple transport modes, spatial scales, and temporal horizons.
These experiments cover forecasting accuracy, computational cost,
joint training versus training one model per mode, and cross-year
generalization.

\subsection{Setup}
\label{sec:experiments-setup}

\paragraph{Task.}
We adopt the standard multi-horizon OD forecasting setting.
Given the past $T$ hours of OD flow $\mathbf{X}_{t-T+1:t} \in \mathbb{R}^{T
\times P \times M}$ across $P{=}Z^2$ OD pairs from $Z$ zones and
$M=6$ transport modes, the model predicts the next $H$ hours $\hat{\mathbf{X}}_{t+1:t+H}$. We evaluate
three forecasting horizons $H \in \{6, 12, 24\}$ with input length
matching the horizon ($T=H$). 
Splits are chronological with a 70/10/20 train/validation/test ratio
over the 2024 calendar year. To test cross-year transfer,
Section~\ref{sec:experiments-crossyear} retrains each model on all of
2024 and evaluates on all of 2025 with the same input/output
configuration.

\paragraph{Baselines.}

We benchmark 14 models in total. The first two are
non-learned baselines. Historical Average (HA) predicts
the mean value for each (hour-of-week, OD pair, mode) combination,
computed on the training portion. Latest Value (LV),
also known as persistence, repeats the most recent observed value
across the forecast horizon.

The remaining 12 are learned baselines, all evaluated in a
joint multi-mode setting. STAEformer~\cite{liu_spatio-temporal_2023},
STID~\cite{shao_spatial-temporal_2022}, AGCRN~\cite{bai_adaptive_2020},
Graph WaveNet~\cite{wu_graph_2019}, and MTGNN~\cite{wu_connecting_2020}
operate on the flattened OD-pair representation.
iTransformer~\cite{liu_itransformer_2024}, SOFTS~\cite{lu_softs_2024},
MTSMixer~\cite{li_mts-mixers_2025}, and TimeMixer~\cite{wang2023timemixer}
are trained on one OD pair at a time and treat the six modes as
interacting variates. MPGCN~\cite{shi_predicting_2020} and
ODCRN~\cite{jiang_countrywide_2021} preserve the $Z{\times}Z$ OD matrix
structure and apply graph convolution along the origin and
destination axes. ODMixer~\cite{liu_odmixer_2025} adopts a
factorized MLP design for fine-grained OD forecasting.

\paragraph{Implementation.}
All learned baselines share the same chronological splits, forecasting
horizons, early-stopping budget, and evaluation metrics. They are run
through BasicTS-compatible configurations and runners~\cite{shao_exploring_2025}.
For MPGCN, ODCRN, and ODMixer, model-specific adapters preserve the
$Z{\times}Z$ OD-matrix layout while writing metrics in the same format.
We report two types of failures. Out-of-memory (OOM) indicates that
a model exceeds the GPU memory budget, while time-exceeded (TE) means
its per-epoch wall-clock time exceeds one hour. All experiments are
conducted on NVIDIA RTX PRO 6000 (97~GB) GPUs. Model-specific wrappers
and other implementation details are provided in
Appendix~\ref{app:benchmark-details}.

\paragraph{Metrics.}
We report three metrics in estimated trip-count units. Mean absolute
error (MAE) serves as the headline metric throughout the paper.
Root mean squared error (RMSE) additionally penalises large individual
prediction errors. Weighted mean absolute percentage error,
$\text{wMAPE} = \sum |\hat{y} - y| / \sum |y|$, is commonly reported
in OD forecasting work~\cite{liu_odmixer_2025,zhang_joint_2026} and
provides a percentage-scale measure that avoids per-observation
division by zero.

\subsection{Multi-Horizon Performance}
\label{sec:experiments-multi-horizon}

Table~\ref{tab:performance} reports overall MAE, RMSE, and wMAPE at
6h, 12h, and 24h. The rightmost columns report 24h computational cost.

At the district level, spatial-interaction models achieve the lowest errors. STAEformer, which applies attention over the flattened OD-pair representation, reports the lowest overall MAE at all three horizons. GNN-based models such as AGCRN, GWNet, and MTGNN generally follow STAEformer. Channel-decomposed transformers exhibit larger errors at 6h and 12h, but move closer to the GNN-based models at 24h. Among non-learned baselines, HA remains competitive with an overall MAE only 6.1\% behind STAEformer at 24h, while LV is uncompetitive at every horizon, with an overall MAE two to three times that of HA.

Across the three horizons, the learned baselines separate into three patterns. STAEformer and STID are nearly horizon-invariant, with district MAE changing only modestly from 6h to 24h. Several spatio-temporal models without temporal embeddings degrade at longer horizons, including AGCRN, GWNet, MTGNN, MPGCN, and ODMixer. By contrast, iTransformer, SOFTS, MTSMixer, and TimeMixer drop from an average MAE of 59.8 at 6h to 28.8 at 24h. In our configurations, STAEformer and STID include explicit time-of-day and day-of-week embeddings, while the pair-wise time-series models use only flow histories. HA's strong performance with just an hour-of-week lookup suggests that recurring daily and weekly rhythms dominate the OD signal. Temporal embeddings encode these rhythms directly, while pair-wise models recover them once the input window covers a full daily cycle.

At the sub-district level, the horizon comparison is limited to the few baselines that complete under the common evaluation protocol. Among completed learned models, ODMixer gives the strongest short-horizon results at 6h and 12h, while STID is strongest at 24h.

For the per-mode breakdown (Table~\ref{tab:performance-h24}), we focus on 24h because it exposes the largest spread between architectures and is the most operationally meaningful for day-ahead planning. HA leads on metro bus, subway, and car, whose hourly profiles are dominated by the weekly cycle. STAEformer leads on local bus, walk, and other. Its 19.6\% reduction in walk MAE alone accounts for most of its 6.1\% lower average MAE relative to HA. On these modes, learned models capture complex patterns that the weekly cycle alone misses.

\begin{table}[t]
  \centering
  \setlength{\tabcolsep}{2.5pt}
  \caption{Multi-horizon performance and computational cost, aggregated
    across modes. Bold marks the best and underline marks the
    second-best per (horizon, metric) within each level. Cost columns
    report parameter count, batch size, per-epoch wall-clock, and
    100-epoch wall-clock estimates at the 24h$\to$24h horizon.}
  \label{tab:performance}
  \resizebox{\textwidth}{!}{%
    \begin{tabular}{l rrr rrr rrr rrrr}
\toprule
 & \multicolumn{3}{c}{6h$\to$6h} & \multicolumn{3}{c}{12h$\to$12h} & \multicolumn{3}{c}{24h$\to$24h} & \multicolumn{4}{c}{Cost (24h)} \\
\cmidrule(lr){2-4} \cmidrule(lr){5-7} \cmidrule(lr){8-10} \cmidrule(lr){11-14}
Model & MAE & RMSE & wMAPE & MAE & RMSE & wMAPE & MAE & RMSE & wMAPE & \#Params & BS & Train/ep & Total \\
\midrule
\multicolumn{14}{l}{\emph{District scale (625 OD pairs)}} \\
\midrule
HA & 22.9 & 112.2 & 10.5\% & \underline{22.9} & 112.2 & \underline{10.5\%} & \underline{22.8} & 111.8 & \underline{10.5\%} & --- & --- & --- & --- \\
LV & 106.6 & 566.8 & 49.0\% & 144.2 & 692.3 & 66.3\% & 139.0 & 678.8 & 63.9\% & --- & --- & --- & --- \\
STAEformer & \textbf{21.8} & \textbf{93.7} & \textbf{10.1\%} & \textbf{22.8} & \textbf{100.1} & \textbf{10.5\%} & \textbf{21.4} & \textbf{92.9} & \textbf{9.8\%} & 2.8M & 8 & 83s & 2.3h \\
STID & 23.4 & 99.7 & 10.8\% & 25.1 & 106.9 & 11.5\% & 24.6 & \underline{111.7} & 11.3\% & 143K & 16 & 3.2s & 0.09h \\
AGCRN & \underline{22.7} & 102.6 & \underline{10.4\%} & 25.2 & 111.2 & 11.6\% & 30.4 & 178.9 & 14.0\% & 780K & 32 & 21s & 0.57h \\
GWNet & 23.8 & \underline{95.9} & 10.9\% & 26.2 & \underline{106.8} & 12.1\% & 27.6 & 118.8 & 12.7\% & 353K & 32 & 31s & 0.86h \\
MTGNN & 23.8 & 101.9 & 11.0\% & 26.7 & 112.4 & 12.3\% & 27.4 & 120.1 & 12.6\% & 1.7M & 32 & 14s & 0.38h \\
iTransformer & 56.7 & 253.8 & 26.1\% & 38.7 & 162.8 & 17.8\% & 25.3 & 119.3 & 11.7\% & 4.8M & 256 & 160s & 4.4h \\
SOFTS & 58.4 & 266.3 & 26.9\% & 42.6 & 180.6 & 19.6\% & 26.5 & 130.1 & 12.2\% & 205K & 256 & 145s & 4.0h \\
MTSMixer & 67.8 & 329.1 & 31.2\% & 51.5 & 243.8 & 23.7\% & 31.9 & 172.7 & 14.7\% & 14K & 256 & 118s & 3.3h \\
TimeMixer & 56.1 & 265.7 & 25.9\% & 51.3 & 253.2 & 23.6\% & 31.6 & 173.0 & 14.5\% & 16K & 256 & 492s & 13.7h \\
ODCRN & 27.8 & 192.6 & 12.8\% & 31.8 & 226.3 & 14.7\% & 30.2 & 226.6 & 13.9\% & 178K & 16 & 233s & 6.5h \\
MPGCN & 34.9 & 232.8 & 16.1\% & 40.5 & 276.7 & 18.7\% & 43.0 & 299.2 & 19.8\% & 75K & 16 & 14s & 0.39h \\
ODMixer & 31.2 & 238.7 & 14.4\% & 37.8 & 269.2 & 17.4\% & 51.9 & 378.0 & 23.9\% & 206K & 16 & 6.8s & 0.19h \\
\midrule
\multicolumn{14}{l}{\emph{Sub-district scale (181{,}476 OD pairs); trained for up to 100 epochs with early-stopping (patience 10)}} \\
\midrule
HA & 0.475 & \textbf{2.81} & 63.5\% & 0.475 & \textbf{2.81} & 63.5\% & 0.475 & \textbf{2.80} & 63.5\% & --- & --- & --- & --- \\
LV & 0.742 & 9.41 & 99.1\% & 0.833 & 11.0 & 111.2\% & 0.820 & 10.9 & 109.6\% & --- & --- & --- & --- \\
STAEformer & --- & --- & --- & --- & --- & --- & --- & --- & --- & 350.0M & 8 & OOM & OOM \\
STID & \underline{0.434} & \underline{5.20} & \underline{58.4\%} & \underline{0.448} & 5.79 & \underline{60.2\%} & \textbf{0.442} & \underline{4.81} & \textbf{59.2\%} & 5.9M & 16 & 1820s & 50.5h \\
AGCRN & --- & --- & --- & --- & --- & --- & --- & --- & --- & 2.6M & 32 & OOM & OOM \\
GWNet & --- & --- & --- & --- & --- & --- & --- & --- & --- & 4.0M & 32 & OOM & OOM \\
MTGNN & --- & --- & --- & --- & --- & --- & --- & --- & --- & 432.8M & 32 & OOM & OOM \\
iTransformer & --- & --- & --- & --- & --- & --- & --- & --- & --- & 4.8M & 256 & TE & TE \\
SOFTS & --- & --- & --- & --- & --- & --- & --- & --- & --- & 205K & 256 & TE & TE \\
MTSMixer & --- & --- & --- & --- & --- & --- & --- & --- & --- & 14K & 256 & TE & TE \\
TimeMixer & --- & --- & --- & --- & --- & --- & --- & --- & --- & 16K & 256 & TE & TE \\
ODCRN & --- & --- & --- & --- & --- & --- & --- & --- & --- & 540K & 16 & OOM & OOM \\
MPGCN & --- & --- & --- & --- & --- & --- & --- & --- & --- & 75K & 16 & OOM & OOM \\
ODMixer & \textbf{0.423} & 5.21 & \textbf{56.6\%} & \textbf{0.434} & \underline{5.54} & \textbf{58.1\%} & 0.485 & 6.96 & 64.9\% & 46.7M & 16 & 540s & 15.0h \\
\bottomrule
\end{tabular}
  }
\end{table}

\begin{table}[t]
  \centering
  \setlength{\tabcolsep}{2.5pt}
  \caption{Per-mode MAE and RMSE at the 24h$\to$24h horizon, district
    level. Bold marks the best and underline marks the second-best per
    (mode, metric).}
  \label{tab:performance-h24}
  \resizebox{\textwidth}{!}{%
    \begin{tabular}{l rr rr rr rr rr rr rr}
\toprule
 & \multicolumn{2}{c}{metro bus} & \multicolumn{2}{c}{local bus} & \multicolumn{2}{c}{subway} & \multicolumn{2}{c}{walk} & \multicolumn{2}{c}{car} & \multicolumn{2}{c}{other} & \multicolumn{2}{c}{Avg} \\
\cmidrule(lr){2-3} \cmidrule(lr){4-5} \cmidrule(lr){6-7} \cmidrule(lr){8-9} \cmidrule(lr){10-11} \cmidrule(lr){12-13} \cmidrule(lr){14-15}
Model & MAE & RMSE & MAE & RMSE & MAE & RMSE & MAE & RMSE & MAE & RMSE & MAE & RMSE & MAE & RMSE \\
\midrule
HA & \textbf{2.00} & \textbf{7.85} & \underline{18.7} & 72.4 & \textbf{32.2} & \underline{73.2} & 35.2 & 233.4 & \textbf{30.5} & \underline{74.7} & \underline{18.1} & 64.9 & \underline{22.8} & 111.8 \\
LV & 6.11 & 34.3 & 104.9 & 417.8 & 231.1 & 466.1 & 232.1 & 1429.7 & 160.0 & 436.3 & 99.8 & 371.0 & 139.0 & 678.8 \\
STAEformer & 2.30 & 11.2 & \textbf{16.5} & \textbf{55.4} & \underline{33.0} & \textbf{72.0} & \textbf{28.3} & \textbf{188.9} & \underline{31.1} & \textbf{69.4} & \textbf{17.0} & \textbf{54.3} & \textbf{21.4} & \textbf{92.9} \\
STID & 2.94 & 13.9 & 19.4 & \underline{65.2} & 37.4 & 82.7 & 33.1 & \underline{231.7} & 35.3 & 78.9 & 19.6 & \underline{60.9} & 24.6 & \underline{111.7} \\
AGCRN & 2.83 & 13.1 & 21.7 & 79.0 & 45.0 & 103.9 & 53.0 & 401.8 & 38.0 & 89.0 & 21.7 & 73.4 & 30.4 & 178.9 \\
GWNet & 2.66 & 12.3 & 21.2 & 70.8 & 41.7 & 84.8 & 41.5 & 248.1 & 37.4 & 81.1 & 20.9 & 64.4 & 27.6 & 118.8 \\
MTGNN & 3.00 & 12.9 & 20.8 & 70.5 & 44.6 & 96.3 & 37.9 & 247.4 & 37.1 & 81.9 & 21.0 & 65.0 & 27.4 & 120.1 \\
iTransformer & \underline{2.07} & \underline{9.03} & 18.7 & 69.4 & 44.5 & 112.6 & \underline{31.9} & 237.9 & 35.2 & 82.0 & 19.6 & 67.7 & 25.3 & 119.3 \\
SOFTS & 2.16 & 9.96 & 19.5 & 74.1 & 47.1 & 125.6 & 32.9 & 259.4 & 36.8 & 87.3 & 20.6 & 72.6 & 26.5 & 130.1 \\
MTSMixer & 3.53 & 14.6 & 23.1 & 87.7 & 57.5 & 154.7 & 42.9 & 361.1 & 41.2 & 98.6 & 23.6 & 83.6 & 31.9 & 172.7 \\
TimeMixer & 2.73 & 13.0 & 22.4 & 90.3 & 57.8 & 160.8 & 41.4 & 356.0 & 41.5 & 104.5 & 23.6 & 87.7 & 31.6 & 173.0 \\
ODCRN & 2.12 & 12.0 & 20.9 & 82.4 & 37.6 & 83.1 & 58.2 & 523.6 & 34.6 & 85.4 & 20.3 & 79.0 & 28.9 & 224.2 \\
MPGCN & 2.55 & 11.7 & 25.1 & 90.5 & 56.8 & 126.7 & 57.4 & 388.1 & 45.0 & 105.2 & 24.4 & 82.6 & 35.2 & 179.3 \\
ODMixer & 5.49 & 26.9 & 29.9 & 122.8 & 64.5 & 147.3 & 79.2 & 610.0 & 52.6 & 141.3 & 31.5 & 120.5 & 43.9 & 272.0 \\
\bottomrule
\end{tabular}
  }
\end{table}

\subsection{Scalability and Computational Cost}
\label{sec:experiments-scalability}

Forecasting OD demand at the sub-district level expands the prediction
target by two orders of magnitude, from 625 to 181{,}476 OD pairs.
We therefore test whether existing models remain computationally feasible
at this scale. The rightmost columns of Table~\ref{tab:performance}
report per-baseline parameter count, batch size, per-epoch wall-clock
time, and estimated 100-epoch wall-clock time at the 24h horizon for
both spatial levels.

At the district level, per-baseline training cost spans nearly
two orders of magnitude. STAEformer achieves the lowest 24h MAE among
learned models, but its 2.8M parameters and over two hours of training
time make it inefficient compared to other baselines. STID, the smallest spatio-temporal baseline,
completes training within minutes and trails STAEformer by only a few
percent in 24h MAE, indicating that heavier
architectures buy limited additional accuracy at this level.

At the sub-district level, the limiting factor depends on the
architecture.
Models that explicitly capture spatial interaction across OD pairs
exceed the available memory, whereas models that treat each OD pair as
an independent sequence fit in memory but exceed the per-epoch time
budget by an order of magnitude. MPGCN and ODCRN, for example, keep
the OD data as dense arrays with shape $[B,T,Z,Z,6]$ and store hidden
activations over both origin and destination axes. At $Z{=}426$ these
activations dominate the small number of stored weights.
Pair-independent transformers avoid dense $Z{\times}Z$ hidden states,
but they train over 181{,}476 OD-pair sequences, which drives the
per-epoch wall-clock past the budget.

\subsection{Cross-Mode Joint Training vs One Model per Mode}
\label{sec:experiments-modesep}

A multi-mode forecasting model can predict the six modes jointly by
setting the output dimension to six, or one model per mode can be
trained separately. The trade-off is intuitive. Joint training exploits
cross-mode signal at the cost of per-mode capacity, while training
one model per mode requires six times the parameters but loses that
shared signal. We examine this trade-off using the four top-performing
models from Table~\ref{tab:performance}. Table~\ref{tab:modesep} compares the two settings at the district 24h horizon.

Training one model per mode outperforms joint training in average MAE
for all four architectures. In our implementation, joint training
uses the six modes as input and output channels but does not explicitly
model dependencies between modes. The
joint setting therefore requires a single shared representation to
predict targets with different scales, sparsity patterns, temporal
profiles, and spatial structures. This suggests that future multi-mode models need mechanisms for
mode-specific patterns and cross-mode signal, instead of just
expanding the output dimension.

\begin{table}[t]
  \centering
  \setlength{\tabcolsep}{2.5pt}
  \caption{Joint training vs one model per mode for four spatio-temporal
    models at the district 24h horizon. Bold marks the lower MAE per
    (model, mode). $\Delta$ is $(\mathrm{Sep}-\mathrm{Joint}) /
    \mathrm{Joint}$.}
  \label{tab:modesep}
  \resizebox{\textwidth}{!}{%
\begin{tabular}{l rrr rrr rrr rrr}
\toprule
 & \multicolumn{3}{c}{STID} & \multicolumn{3}{c}{AGCRN}
 & \multicolumn{3}{c}{STAEformer} & \multicolumn{3}{c}{MTGNN} \\
\cmidrule(lr){2-4} \cmidrule(lr){5-7} \cmidrule(lr){8-10}
\cmidrule(lr){11-13}
Mode & Joint & Sep & $\Delta$ & Joint & Sep & $\Delta$
     & Joint & Sep & $\Delta$ & Joint & Sep & $\Delta$ \\
\midrule
metro\_bus & 2.94 & \textbf{1.79} & $-39.0\%$ & 2.83 & \textbf{2.05} & $-27.8\%$ & 2.30 & \textbf{1.81} & $-21.0\%$ & 3.00 & \textbf{1.86} & $-38.1\%$ \\
local\_bus & 19.36 & \textbf{15.85} & $-18.1\%$ & 21.71 & \textbf{18.74} & $-13.7\%$ & 16.51 & \textbf{16.01} & $-3.0\%$ & 20.80 & \textbf{17.52} & $-15.7\%$ \\
subway & 37.43 & \textbf{30.98} & $-17.2\%$ & 44.98 & \textbf{41.17} & $-8.5\%$ & 32.95 & \textbf{29.37} & $-10.9\%$ & 44.56 & \textbf{34.14} & $-23.4\%$ \\
walk & 33.14 & \textbf{27.77} & $-16.2\%$ & 53.02 & \textbf{43.50} & $-18.0\%$ & \textbf{28.33} & 29.80 & $+5.2\%$ & 37.87 & \textbf{35.50} & $-6.3\%$ \\
car & 35.32 & \textbf{29.38} & $-16.8\%$ & 37.96 & \textbf{32.27} & $-15.0\%$ & 31.12 & \textbf{27.90} & $-10.3\%$ & 37.13 & \textbf{31.85} & $-14.2\%$ \\
other & 19.58 & \textbf{16.27} & $-16.9\%$ & 21.72 & \textbf{19.07} & $-12.2\%$ & 16.98 & \textbf{16.32} & $-3.9\%$ & \textbf{21.01} & 22.60 & $+7.6\%$ \\
\midrule
Avg & 24.63 & \textbf{20.34} & $\mathbf{-17.4\%}$ & 30.37 & \textbf{26.13} & $\mathbf{-14.0\%}$ & 21.37 & \textbf{20.20} & $\mathbf{-5.4\%}$ & 27.40 & \textbf{23.91} & $\mathbf{-12.7\%}$ \\
\bottomrule
\end{tabular}
  }
\end{table}

\subsection{Cross-Year Generalization}
\label{sec:experiments-crossyear}

Cross-year transfer is rarely evaluated in OD forecasting because
public datasets seldom span more than one calendar year. The
pronounced month-to-month variability of \dataset
(Figure~\ref{fig:monthly}) makes its 2025 calendar a natural
robustness test for models trained on 2024.
To test whether models trained on 2024 transfer to 2025, we
retrain five architectures (STAEformer, AGCRN,
Graph WaveNet, MTGNN, and iTransformer) on the full 2024 calendar
and evaluate on the full 2025. Table~\ref{tab:crossyear} reports
overall and per-season MAE at the district 24h horizon.

All five models have higher MAE in cross-year evaluation, with
iTransformer showing the smallest increase. The seasonal
ordering is consistent across architectures, with spring being the
hardest and summer the easiest.

\paragraph{Contextual extensions.}
We explore how three contextual signals released with the dataset
(holiday calendar, hourly rainfall, per-district POI counts) change
forecasting accuracy. Each of the three signals lowers 24h MAE by 0.49 to 1.39 when added to STID, but their effects differ at shorter horizons: the holiday embedding raises MAE at 6h and 12h, rainfall raises MAE at 6h but lowers it at 12h, and POI counts stay neutral or slightly helpful at every horizon. Appendix~\ref{app:context-predictive} gives the full table and setup.

\begin{table}[t]
  \centering
  \setlength{\tabcolsep}{2.5pt}
  \caption{Cross-year generalization at the district 24h horizon
    (train 2024, test 2025), broken down by season. Bold marks the
    easiest season per model. $\Delta\%$ is the in-year vs cross-year
    MAE change.}
  \label{tab:crossyear}
\begin{tabular}{l rr rrrr r}
\toprule
Model        & 2024 test & 2025 overall & Spring & Summer & Fall & Winter & $\Delta$ \\
\midrule
STAEformer & 21.37 & 25.83 & 28.29 & \textbf{23.70} & 25.75 & 25.56 & $+20.89\%$ \\
AGCRN & 30.37 & 35.77 & 37.33 & \textbf{32.84} & 36.93 & 36.00 & $+17.78\%$ \\
GWNet & 27.55 & 33.47 & 35.21 & \textbf{30.60} & 33.84 & 34.29 & $+21.49\%$ \\
MTGNN & 27.40 & 32.35 & 34.63 & \textbf{29.95} & 31.93 & 32.89 & $+18.07\%$ \\
iTransformer & 25.33 & 28.08 & 29.87 & \textbf{26.65} & 28.35 & 27.42 & $+10.86\%$ \\
\bottomrule
\end{tabular}
%
\end{table}


\section{Conclusion}
\label{sec:conclusion}

We introduced \dataset, a city-scale multi-mode OD forecasting benchmark
for Seoul that combines six-mode hourly flows with trip
attributes and contextual information. This benchmark highlights two
central challenges for current OD forecasting models in city-scale
multi-mode settings. Spatio-temporal models that
perform well at district level do not scale to sub-district
forecasting, and naive joint multi-mode training does not
consistently improve accuracy.
Future model design should scale to all origin-destination pairs at
fine spatial resolution and use cross-mode information more directly.
\dataset provides a reproducible testbed for developing and evaluating
such models.


\section*{Reproducibility statement}

All learned baselines are run through the same BasicTS-style training
and evaluation pipeline~\cite{shao_exploring_2025}. The project
repository releases the dataset builders, model configurations,
BasicTS-compatible adapters, and training and evaluation scripts used
in the benchmark. Appendix~\ref{app:benchmark-details} documents the
benchmark protocol and model-specific implementation details. The
dataset release provides the OD flow parquet files and metadata needed
to build the benchmark tensors used for Table~\ref{tab:performance}.



\printbibliography

\newpage
\appendix

\section{Additional Dataset Analysis}
\label{app:additional-dataset-analysis}

\begin{figure}[t]
  \centering
  \begin{subfigure}[t]{0.24\textwidth}
    \includegraphics[width=\textwidth]{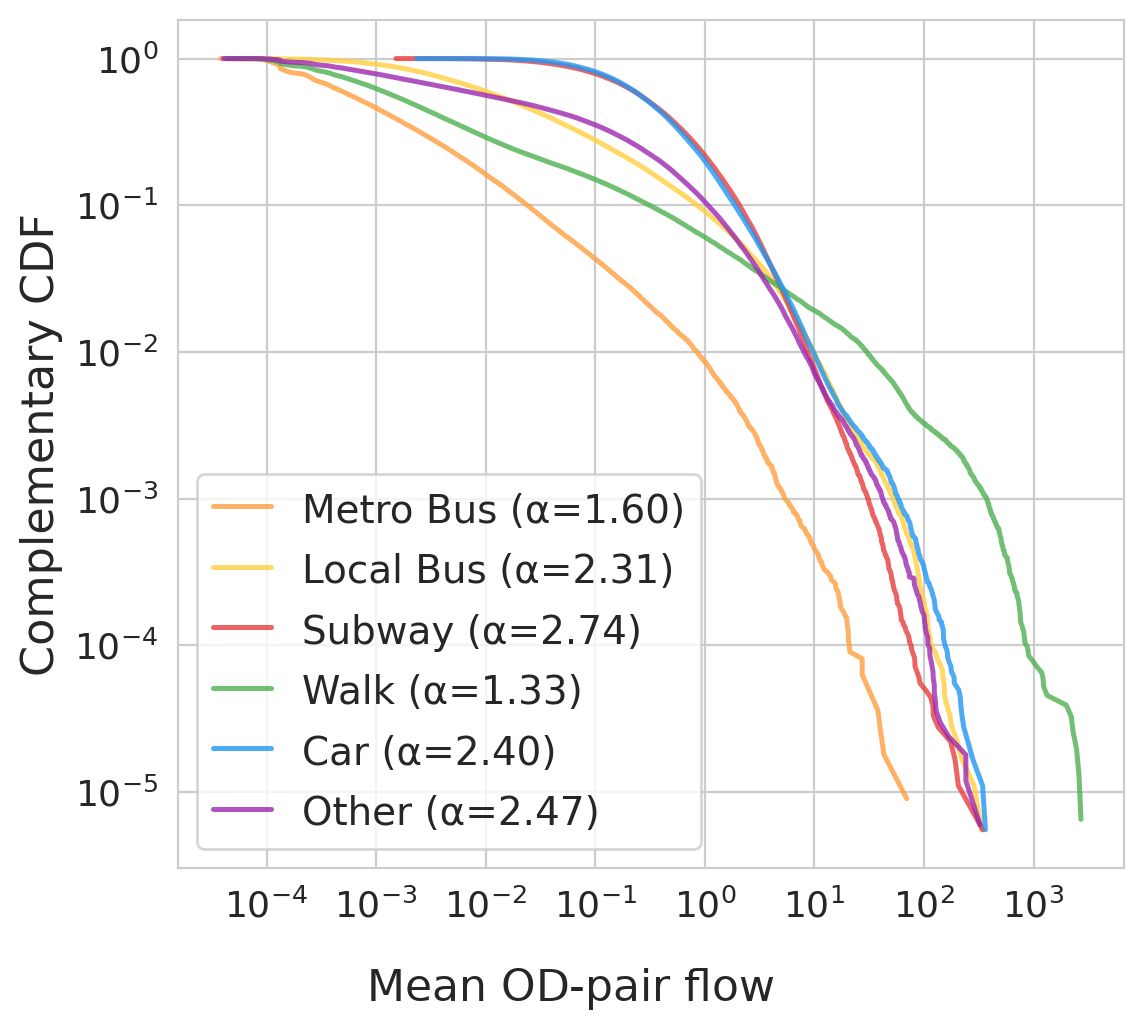}
    \caption{Heavy-tailed CCDF}
    \label{fig:powerlaw}
  \end{subfigure}\hfill
  \begin{subfigure}[t]{0.24\textwidth}
    \includegraphics[width=\textwidth]{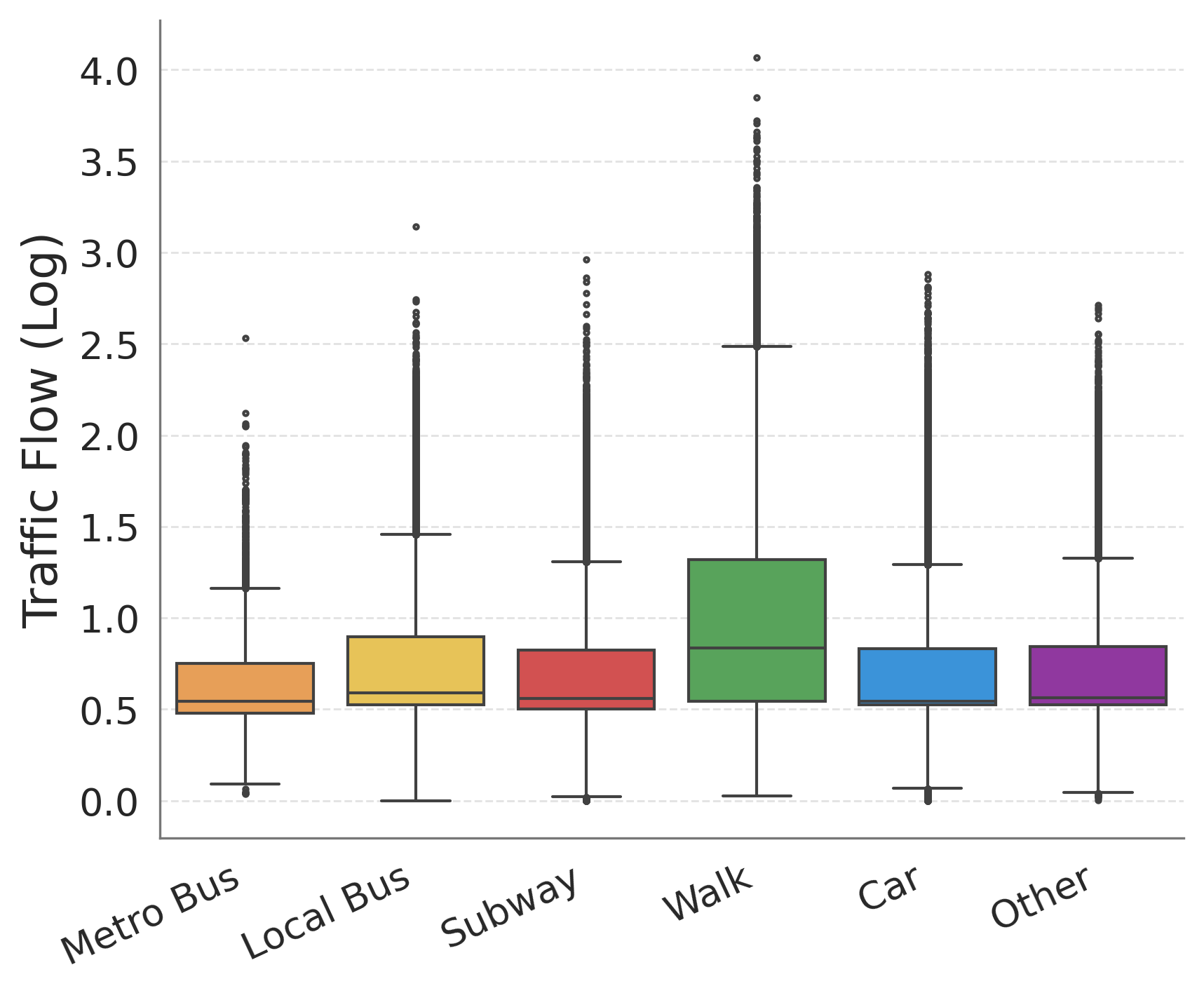}
    \caption{Flow by mode}
    \label{fig:flow-mode}
  \end{subfigure}\hfill
  \begin{subfigure}[t]{0.24\textwidth}
    \includegraphics[width=\textwidth]{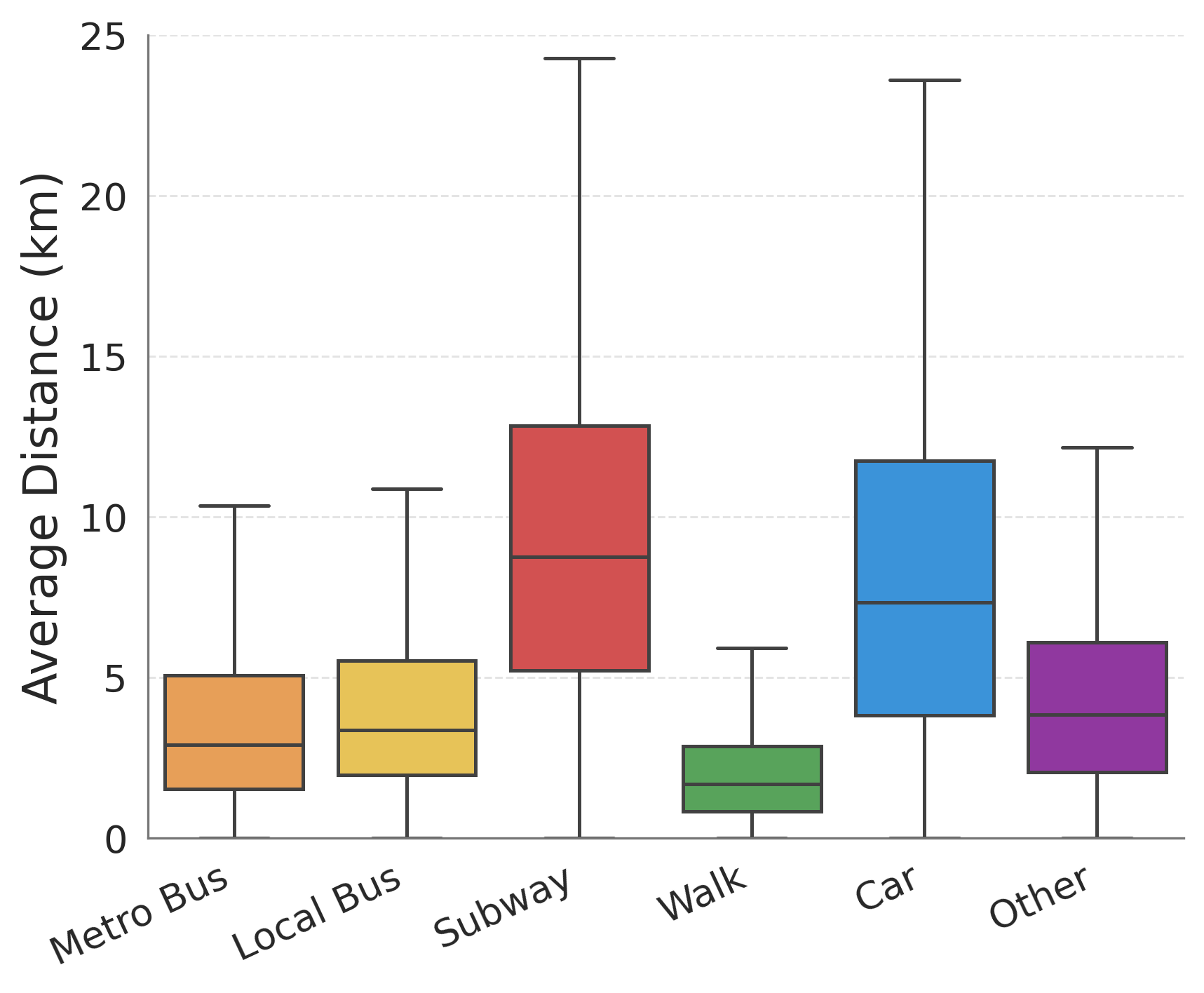}
    \caption{Distance by mode}
    \label{fig:mode-distance}
  \end{subfigure}\hfill
  \begin{subfigure}[t]{0.24\textwidth}
    \includegraphics[width=\textwidth]{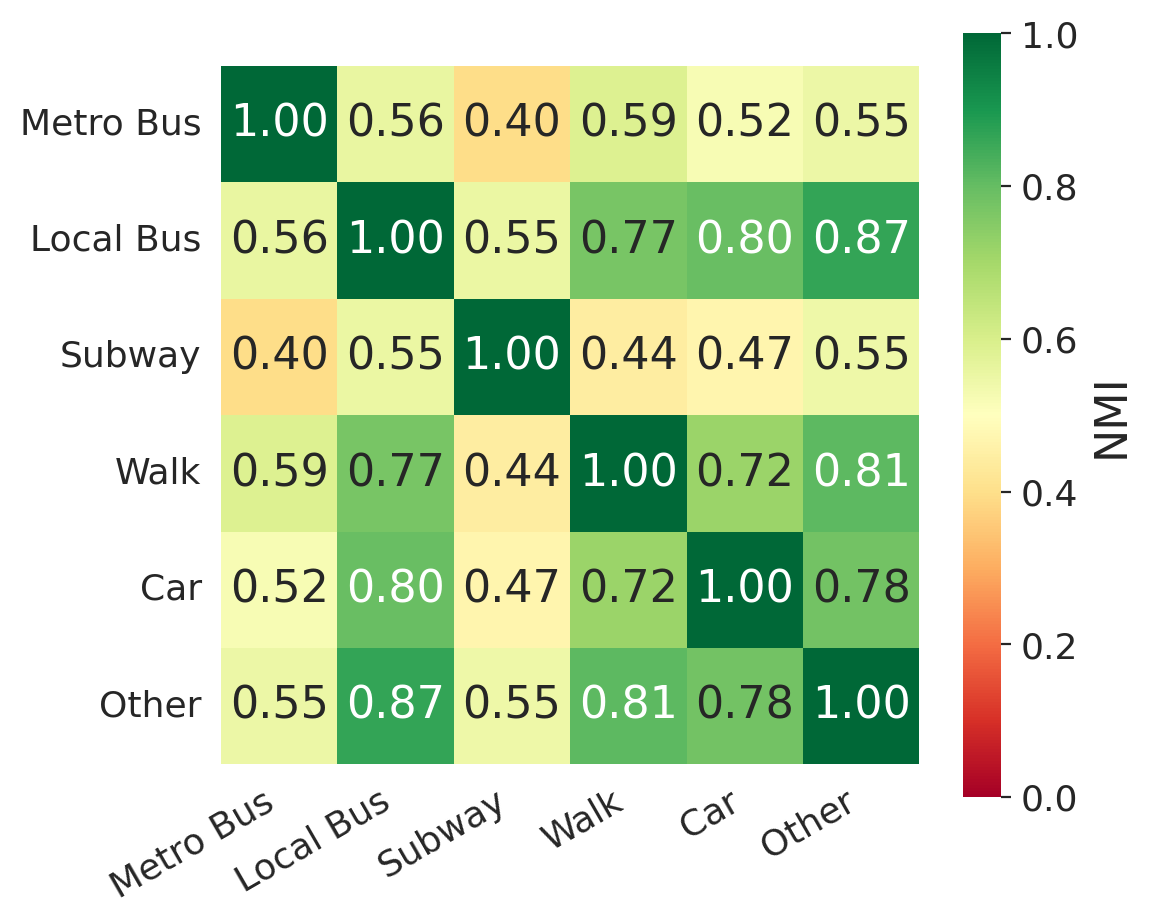}
    \caption{Community NMI}
    \label{fig:nmi}
  \end{subfigure}

  \caption{Additional descriptive views of multi-mode OD structure.
    (a)~Complementary cumulative distribution of OD-pair flows per mode.
    (b)~Distribution of OD-record flow by mode.
    (c)~Distribution of average OD-record distance by mode.
    (d)~Normalised mutual information between per-mode community
    assignments.}
  \label{fig:additional-multimode}
\end{figure}

Figure~\ref{fig:additional-multimode} complements
Section~\ref{sec:dataset-analysis} with four descriptive views of the
OD data. Mode-specific flow distributions are heavy-tailed,
and trip volumes and distances vary by mode, suggesting
different hub sizes and interaction ranges per mode. We examine these
differences at the network level by comparing Louvain community
partitions across modes with normalised mutual information (NMI).
Figure~\ref{fig:nmi}
shows generally low NMI values, suggesting that spatial community
assignments differ across modes.

\section{Contextual Signal Analysis}
\label{app:context-analysis}

\dataset bundles three contextual signals (Section~\ref{sec:dataset-context})
to support context-aware extensions without re-collecting data. We
describe each signal (Calendar, Rainfall, POI) and then evaluate predictive utility and error stratification in
Sections~\ref{app:context-predictive} and~\ref{app:context-error}.

\textbf{Calendar.} Table~\ref{tab:calendar-mode} compares total daily
trip volume and per-mode share across weekdays, non-holiday weekends,
and Korean public holidays in 2024. Total demand drops 18\% on
weekends and 29\% on holidays. The subway share falls by 4.1\,pp on
weekends and 5.5\,pp on holidays. Walking ($+$3.3\,pp on holidays) and
car ($+$3.0\,pp) take over most of this share, consistent with a shift
from commuting to local leisure trips.

\begin{table}[t]
  \centering
  \footnotesize
  \setlength{\tabcolsep}{5pt}
  \caption{Mode share (\%) and total daily trip volume by day type
    for 2024 at the district scale. \textit{Holiday} counts Korean
    public holidays. \textit{Weekend} excludes days flagged as
    holidays. Mode shares sum to 100\% per row.}
  \label{tab:calendar-mode}
  \begin{tabular}{l rr rrrrrr}
    \toprule
    Day type & $n$ days & Trips/day & metro bus & local bus & subway & walk & car & other \\
    \midrule
    Weekday           & 246 & 20.64\,M & 0.61 & 12.48 & 21.73 & 33.06 & 20.66 & 11.45 \\
    Weekend           & 101 & 16.87\,M & 0.59 & 12.80 & 17.63 & 35.06 & 22.80 & 11.12 \\
    Holiday           &  19 & 14.71\,M & 0.59 & 12.58 & 16.24 & 36.35 & 23.61 & 10.63 \\
    \midrule
    \multicolumn{3}{l}{\textit{Change vs weekday (pp)}}
                            & metro bus & local bus & subway & walk & car & other \\
    \midrule
    Weekend           & --- &  $-$18\%   & $-$0.02 & $+$0.32 & $-$4.10 & $+$1.99 & $+$2.14 & $-$0.33 \\
    Holiday           & --- &  $-$29\%   & $-$0.02 & $+$0.10 & $-$5.50 & $+$3.29 & $+$2.95 & $-$0.82 \\
    \bottomrule
  \end{tabular}
\end{table}

\textbf{Rainfall.} Table~\ref{tab:rain-mode} reports origin district-hour
mode shares by rainfall intensity. Trips on dry district-hours are split
20.5\% subway, 33.7\% walk, 21.3\% car, with the remainder in bus and
other modes. Heavy-rain cells ($>$10\,mm/hour) account for 0.30\% of
district-hours. In these cells, walk share drops to 30.6\% and subway
share rises to 22.4\%. Rainfall shifts mode shares modestly, so we release
it as an auxiliary feature rather than the main target.

\begin{table}[t]
  \centering
  \footnotesize
  \setlength{\tabcolsep}{5pt}
  \caption{Origin district-hour mode share (\%) by rainfall intensity,
    district scale, 2024. Mode shares sum to 100\% per row.}
  \label{tab:rain-mode}
  \begin{tabular}{lrr rrrrrr}
    \toprule
    Condition & $n$ cells & Share & metro bus & local bus & subway & walk & car & other \\
    \midrule
    Dry (0\,mm)              & 206{,}895 & 94.21\% & 0.60 & 12.55 & 20.49 & 33.74 & 21.30 & 11.33 \\
    Trace ($<$0.5\,mm)         &   3{,}012 &  1.37\% & 0.67 & 12.38 & 21.22 & 32.44 & 21.75 & 11.54 \\
    Light (0.5--2\,mm)         &   5{,}741 &  2.61\% & 0.63 & 12.74 & 21.09 & 33.13 & 20.98 & 11.43 \\
    Moderate (2--10\,mm)       &   3{,}283 &  1.49\% & 0.62 & 13.04 & 20.88 & 32.43 & 21.44 & 11.58 \\
    Heavy ($>$10\,mm)          &     669   &  0.30\% & 0.62 & 13.42 & 22.42 & 30.60 & 21.06 & 11.87 \\
    \bottomrule
  \end{tabular}
\end{table}

\textbf{POI categories.} Per-sub-district POI counts cover 20 categories,
which we group into four broad groups for presentation.
Table~\ref{tab:poi-families} lists the categories in each group. To
check whether these counts predict the mode mix at each destination,
we correlate each category's count with the share of arrivals using
each mode in 2024 (Table~\ref{tab:poi-corr}). For each (POI, mode)
pair, we report raw Pearson correlation, Spearman rank correlation,
Pearson correlation on POI density, and partial Pearson correlation
controlling for sub-district area. FDR-adjusted $q$-values are
computed across all 120 POI--mode pairs, and the table reports entries
with raw $|r|{\geq}0.20$ and $q{<}0.001$.

\begin{table}[t]
  \centering
  \footnotesize
  \setlength{\tabcolsep}{4pt}
  \caption{POI count features grouped into four broad groups for presentation.}
  \label{tab:poi-families}
  \begin{tabular}{p{0.27\linewidth}p{0.65\linewidth}}
    \toprule
    Group & Categories \\
    \midrule
    Transport infrastructure & airport, rail station, bus terminal, subway station, bus stop \\
    Commerce and leisure & attraction, bank, department store, supermarket, theater, hotel \\
    Public and health services & government office, general hospital, clinic, pharmacy \\
    Education & kindergarten, elementary school, middle school, high school, university \\
    \bottomrule
  \end{tabular}
\end{table}

Subway infrastructure gives the clearest signal. Subway-station count
correlates with subway arrival share at $r{=}{+}0.458$ and remains
stable under the rank, density, and area-controlled variants. Bus-stop
count tracks local-bus share, while commerce and leisure categories
such as banks and theaters track subway share. Education categories,
especially kindergarten, correlate negatively with subway share.

For bus stops and walk share, the raw count correlation is negative
($r{=}{-}0.284$), but it changes to $r{=}{+}0.184$ under density and
weakens to $r{=}{-}0.097$ after controlling for sub-district area.
Sub-district size explains the raw association because large
sub-districts can contain many bus stops while having low walk share
due to longer typical trip distances.

\begin{table}[t]
  \centering
  \footnotesize
  \setlength{\tabcolsep}{4pt}
  \caption{Correlation between per-sub-district POI count and
    destination mode share. We report four correlation variants per
    (POI, mode) pair --- Pearson $r$ on counts, Spearman $r_s$,
    Pearson $r$ on per-km$^2$ density, and partial $r$ controlling
    for sub-district area.
    Rows have raw $|r|{\geq}0.20$ and BH-adjusted $q{<}0.001$.}
  \label{tab:poi-corr}
  \begin{tabular}{ll rrrr}
    \toprule
    POI & Mode & $r$ (count) & $r_s$ & $r$ (density) & $r$ (partial $|$ area) \\
    \midrule
    Subway station  & subway     & $+$0.458 & $+$0.442 & $+$0.438 & $+$0.475 \\
    Bank            & subway     & $+$0.356 & $+$0.314 & $+$0.430 & $+$0.369 \\
    Theater         & subway     & $+$0.345 & $+$0.298 & $+$0.332 & $+$0.347 \\
    Subway station  & local bus  & $-$0.284 & $-$0.324 & $-$0.256 & $-$0.286 \\
    Bus stop        & local bus  & $+$0.276 & $+$0.324 & $+$0.318 & $+$0.338 \\
    Kindergarten    & subway     & $-$0.283 & $-$0.307 & $-$0.257 & $-$0.283 \\
    Attraction      & subway     & $+$0.277 & $+$0.221 & $+$0.307 & $+$0.307 \\
    Gov office      & subway     & $+$0.259 & $+$0.190 & $+$0.295 & $+$0.276 \\
    Bank            & metro bus  & $+$0.239 & $+$0.367 & $+$0.215 & $+$0.222 \\
    Pharmacy        & subway     & $+$0.239 & $+$0.209 & $+$0.222 & $+$0.245 \\
    Bank            & local bus  & $-$0.221 & $-$0.233 & $-$0.223 & $-$0.220 \\
    Bus stop        & walk       & $-$0.284 & $-$0.219 & $+$0.184 & $-$0.097 \\
    \bottomrule
  \end{tabular}
\end{table}

\subsection{Predictive utility of contextual signals}
\label{app:context-predictive}

To evaluate whether contextual signals improve forecasts on top of
mode-flow history, we extend STID with three variants and measure the
change in test MAE on the district-scale 2024 split. STID's
embedding-and-MLP design lets each variant add a single lookup or
projection in isolation, so per-signal effects are easy to compare. \textit{Holiday} adds a 3-state day-type
embedding (workday / weekend / holiday) read from the input and
prediction windows. \textit{POI} attaches a static per-pair feature
built from 13 district-level POI categories. The origin and
destination count vectors are concatenated, log-transformed, z-scored,
and projected to a 32-d embedding shared across timesteps.
\textit{Rainfall} adds two dynamic channels, hourly rainfall at the
origin and destination districts from the Seoul AWS network,
log1p-transformed and z-scored. Each dynamic channel passes through a
separate $1{\times}1$ convolution to keep its projection independent
from the mode-flow channels.

\textbf{Setup.} We follow the benchmark protocol described in
Appendix~\ref{app:benchmark-details}. All STID variants share an embed
dim of 32, 3 encoder layers, learning rate $2{\cdot}10^{-3}$, and
MultiStepLR with milestones $\{1, 50, 80\}$ ($\gamma{=}0.5$). We report
the test MAE produced by the best validation checkpoint per (variant,
horizon).

\begin{table}[t]
  \centering
  \footnotesize
  \setlength{\tabcolsep}{6pt}
  \caption{Test MAE on the 2024 district-scale dataset when STID is extended with
    each contextual signal in isolation. Lower is better, bold marks
    the best per horizon, and $\Delta$ is the change against
    \textit{STID base}.}
  \label{tab:context-mae}
  \begin{tabular}{l rrr rrr}
    \toprule
    Variant   & MAE @ 6 & MAE @ 12 & MAE @ 24 & $\Delta$ @ 6 & $\Delta$ @ 12 & $\Delta$ @ 24 \\
    \midrule
    STID base & 23.41 & 25.07 & 24.63 & ---     & ---     & ---     \\
    + Holiday & 23.81 & 25.55 & \textbf{23.24} & $+$0.40 & $+$0.48 & $-$1.39 \\
    + POI     & \textbf{22.97} & 25.35 & 24.14 & $-$0.44 & $+$0.28 & $-$0.49 \\
    + Rainfall & 24.75 & \textbf{24.37} & 24.14 & $+$1.34 & $-$0.70 & $-$0.49 \\
    \bottomrule
  \end{tabular}
\end{table}

All three contextual signals lower 24-hour MAE. Holiday lowers MAE by
1.39, POI by 0.49, and Rainfall by 0.49. At shorter horizons the three
signals behave differently. Holiday hurts $h{=}6$ and $h{=}12$
($+$0.40, $+$0.48 MAE), so it helps only when the prediction window
spans multiple days. Rainfall reverses sign with horizon, going from
$+$1.34 MAE at $h{=}6$ to $-$0.70 at $h{=}12$. POI helps at $h{=}6$ and $h{=}24$ but costs $+$0.28 MAE at $h{=}12$, so no variant improves MAE at every horizon.

We make the variants, configs, and build scripts available so users
can reproduce the table or run multi-seed experiments.

\subsection{Error stratification by contextual signals}
\label{app:context-error}

\begin{figure}[t]
  \centering
  \includegraphics[width=\linewidth]{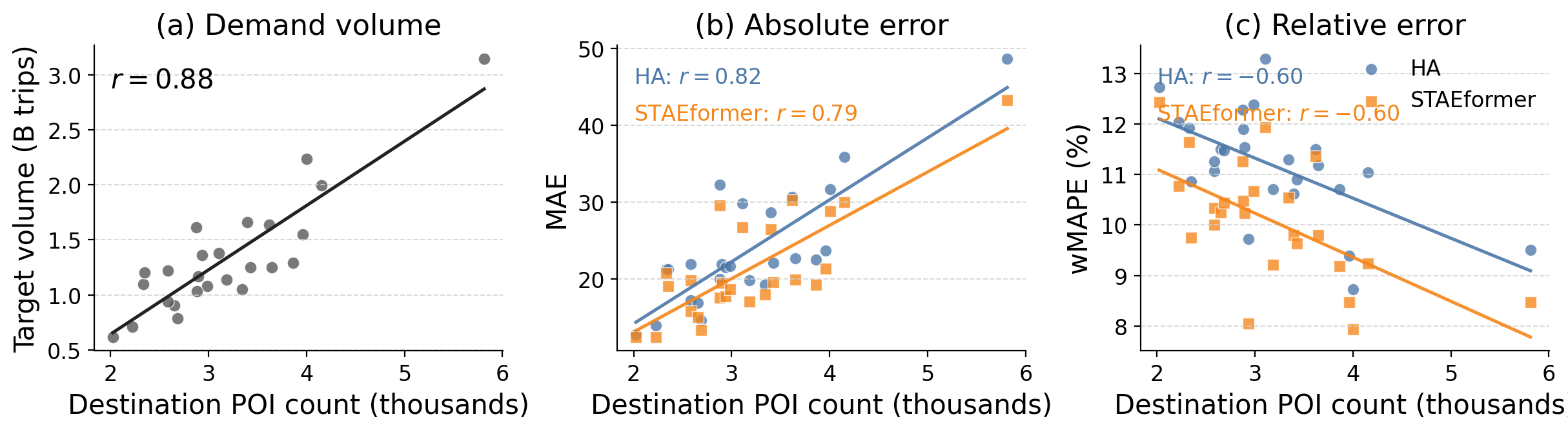}
  \caption{Error stratification by destination-district POI count.
    (a)~Districts with more POIs have larger target volume.
    (b)~Absolute MAE increases with POI count for HA and STAEformer.
    (c)~Relative error decreases with POI count, indicating that
    high-volume districts can have larger absolute errors but smaller
    percentage errors. Pearson correlations are shown in each panel.}
  \label{fig:context-error}
\end{figure}

These contextual signals let users diagnose model behavior after
training. As one example, we aggregate 24-hour district-scale test
errors by destination district and compare them with the district's
POI count. Figure~\ref{fig:context-error} shows that districts with
more POIs have higher target volume and larger absolute errors for
both HA and STAEformer. These districts have lower wMAPE, showing
that absolute error and relative error can tell different stories about
the same urban contexts.

This POI-error comparison is diagnostic rather than causal. Districts
with more POIs concentrate more demand, so the plot should not be read
as POIs causing model error. The point is that \dataset's
contextual signals let users inspect where forecasting errors
concentrate across urban environments.

\section{Benchmark Implementation Details}
\label{app:benchmark-details}

All twelve learned baselines use the same 70/10/20 chronological train/val/test split,
a maximum of 100 epochs, Adam optimization, validation and test evaluation after every
epoch, and validation MAE for checkpoint selection. Where early stopping is enabled,
the patience is 10 epochs. Learning rates, batch sizes, weight decay, gradient clipping, and
learning-rate schedules remain model-specific. Each baseline follows
its source configuration or framework default, with level-specific
changes recorded in the released configs. All learned models predict
all horizons jointly with multimodal output ($M{=}6$) and report MAE,
RMSE, and wMAPE after inverse scaling. Throughout this appendix, $Z$ denotes the number of zones and
$P{=}Z^2$ the number of OD pairs.

\subsection{Spatio-temporal forecasting baselines}
Nine spatio-temporal baselines (STAEformer, STID, AGCRN, Graph WaveNet, MTGNN,
iTransformer, SOFTS, MTSMixer, TimeMixer) run within BasicTS~\cite{shao_exploring_2025}
with masked-MAE loss. STAEformer, STID, AGCRN, Graph WaveNet, and MTGNN use the
BasicTS Z-score scaler, while iTransformer, SOFTS, MTSMixer, and TimeMixer use a
global Z-score variant for pair-wise time-series inputs. In all cases, scaling is
applied only to the six mode-flow channels; calendar, holiday, distance, time, and
rainfall channels are passed through unchanged.

Graph WaveNet and MTGNN expose an \texttt{out\_dim} argument meant for single-channel
multi-step output. We wrap them with an adapter that sets $\texttt{out\_dim}{=}H{\times}6$
and reshapes the flat output to $[B, H, P, 6]$.

\subsection{OD flow forecasting baselines}
Three OD flow forecasting baselines (MPGCN, ODCRN, ODMixer) preserve the $Z{\times}Z$ OD
matrix structure rather than flattening to OD-pair sequences. They use
BasicTS-compatible adapters and MAE loss.

MPGCN and ODCRN use a training adapter that feeds the matrix in $[B, T, Z, Z, 6]$ form
rather than BasicTS's $[B, T, P, 6]$ flattening, with input and output projections that
lift the six-mode channel dimension into and out of the model's hidden width. ODMixer
uses mode channels folded along the temporal axis and receives the previous-day window
used by its trend-learning branch.

\subsection{Parameter-free baselines}
HistoricalAverage and LatestValue share the BasicTS preprocessing and runner code with
the learned baselines. HistoricalAverage builds a 168-bucket hour-of-week table over the
training portion in normalized flow space and maps each test timestamp to its bucket via
the dataset's tod/dow channels. LatestValue copies the last input frame across the
prediction horizon.

\section{Dataset Documentation for \dataset}
\label{app:datasheet}

\subsection{Provenance}
\dataset is built from the Seoul Metropolitan Government's ``Seoul
Metropolitan Area Living Mobility Data,'' jointly produced with KT
(formerly Korea Telecom). We downloaded ZIP-compressed CSV files from
January 2023 through December 2025 from the Seoul Open Data Plaza and
converted them into released parquet files and the benchmark format in
Section~\ref{sec:dataset-source}.

\subsection{Composition}
Each record contains the population-scaled number of trips from
origin zone $i$ to destination zone $j$ during hour $t$ by mode $m$.
\dataset uses two Seoul zone systems, 25 districts and 426
sub-districts, and six intra-urban modes, metro bus, local bus,
subway, walk, car, and other.

For 2024, the district representation contains
$8{,}784\times625\times6 \approx 3.3\times10^{7}$ entries and the
sub-district representation contains
$8{,}784\times181{,}476\times6 \approx 9.6\times10^{9}$ entries. The
complete dataset covers 2023, 2024, and 2025 at both spatial scales,
with aligned calendar, rainfall, and POI signals described in
Section~\ref{sec:dataset-context}. The released parquet files store
observed OD records sparsely and do not include rows for absent
OD-pair, mode, and hour combinations. When building benchmark tensors,
these absent cells are filled with zeros, yielding the dense shapes
above. Two sub-district boundary changes, a 2023-02 rename and a 2025-08
split, are reconciled in the OD-flow records to maintain stable zone
identities.

\dataset contains aggregated administrative-zone OD records only. It
contains no raw subscriber records, trajectories, or device identifiers
and does not support individual-level tracking or re-identification.

Recommended splits are the chronological 70/10/20 split within 2024
(Section~\ref{sec:experiments-setup}) and the cross-year setup that
trains on 2024 and tests on 2025
(Section~\ref{sec:experiments-crossyear}).

\subsection{Upstream Generation}
The public source documentation describes how KT LTE/5G signalling logs are
cleaned, how stays are separated from movements using a 30-minute
criterion, and how mode labels are assigned. Intercity modes are
inferred from the origin--destination zone pair. Subway and bus trips
are assigned by matching inferred paths to published transit networks.
Remaining trips are classified as car or walk by speed and distance,
and unmatched trips are assigned to ``other.''

The original source scales each KT subscriber's trips up to a
population-level estimate. The scaling uses KT market share, LTE/5G
subscriber share, phone-on ratio, and a gender-age adjustment against
the national resident-registration population. \dataset uses the
published OD flows as its flow target. These corrections reduce
estimation error from relying on one carrier's subscriber base but
cannot remove all residual error from phone-use behaviour, visitors,
or under-represented subscriber groups.

\subsection{Our Preprocessing}
Four preprocessing steps convert the original source records into
\dataset.
\begin{enumerate}[leftmargin=*,itemsep=1pt,topsep=2pt]
  \item \emph{Spatial filtering.} Keep only Seoul-to-Seoul OD records.
  \item \emph{Mode filtering.} Drop intercity modes and keep the six
  intra-urban modes.
  \item \emph{Code reconciliation.} Map the 2023-02 rename and 2025-08
  split to stable zone identities.
  \item \emph{Temporal unification.} Drop sub-hour timestamps and
  re-aggregate records to a 1\,h grid.
\end{enumerate}
Timestamps are aligned to Asia/Seoul, which has no daylight-saving
discontinuities. Calendar covariates are normalised to $[0,1)$.
The project repository provides data builders for converting the
released parquet files into benchmark tensors.

\subsection{Uses and Limitations}
\dataset is intended for multi-horizon OD forecasting, comparing joint
training with one model per mode, cross-year generalisation, and context-aware
modelling with the bundled rainfall, calendar, and POI signals.
Individual-level tracking or re-identification is out of scope.

\dataset is a single-city benchmark. Strong performance here does not
by itself establish transfer to other cities, operators, or zoning
systems. The OD flows are upstream mobile-network estimates rather than
direct counts of all trips. Although the source applies population
scaling, estimation error may remain for visitors, phone-use patterns,
or under-represented groups.

\subsection{Distribution}
During review, the anonymised dataset and code are mirrored at
\url{https://anonymous.4open.science/r/SeoulMMOD}. The public dataset
will be hosted on Kaggle. The underlying Seoul data are distributed
under the Korea Open Government License (KOGL) Type~1. The processed
\dataset release is distributed under CC BY-NC 4.0, with attribution
to the original Seoul Open Data Plaza source.


\section*{NeurIPS Paper Checklist}

\begin{enumerate}

\item {\bf Claims}
    \item[] Question: Do the main claims made in the abstract and introduction accurately reflect the paper's contributions and scope?
    \item[] Answer: \answerYes{}
    \item[] Justification: The abstract and introduction state the dataset
    contribution, benchmark scope, and main empirical findings. The dataset
    description and experiments provide the corresponding evidence.

\item {\bf Limitations}
    \item[] Question: Does the paper discuss the limitations of the work performed by the authors?
    \item[] Answer: \answerYes{}
    \item[] Justification: The paper discusses the limits of the benchmark,
    including its single-city scope, possible bias in the upstream
    mobile-network estimates, and the fact that it does not establish
    transfer to other cities, operators, or zoning systems.

\item {\bf Theory assumptions and proofs}
    \item[] Question: For each theoretical result, does the paper provide the full set of assumptions and a complete (and correct) proof?
    \item[] Answer: \answerNA{}
    \item[] Justification: The paper does not introduce theoretical results,
    theorems, or proofs.

    \item {\bf Experimental result reproducibility}
    \item[] Question: Does the paper fully disclose all the information needed to reproduce the main experimental results of the paper to the extent that it affects the main claims and/or conclusions of the paper (regardless of whether the code and data are provided or not)?
    \item[] Answer: \answerYes{}
    \item[] Justification: We specify the experimental protocol and release
    code, data, scripts, and configurations for reproducing the benchmark
    results.

\item {\bf Open access to data and code}
    \item[] Question: Does the paper provide open access to the data and code, with sufficient instructions to faithfully reproduce the main experimental results, as described in supplemental material?
    \item[] Answer: \answerYes{}
    \item[] Justification: We provide dataset and code URLs. The release
    includes instructions, preprocessing scripts, model configurations,
    and training and evaluation commands.

\item {\bf Experimental setting/details}
    \item[] Question: Does the paper specify all the training and test details (e.g., data splits, hyperparameters, how they were chosen, type of optimizer) necessary to understand the results?
    \item[] Answer: \answerYes{}
    \item[] Justification: The paper specifies the data splits, horizons,
    baselines, metrics, and training protocol. Full model configurations and
    hyperparameters are provided with the released code.

\item {\bf Experiment statistical significance}
    \item[] Question: Does the paper report error bars suitably and correctly defined or other appropriate information about the statistical significance of the experiments?
    \item[] Answer: \answerNo{}
    \item[] Justification: We report one run for each model and evaluation
    setting, and do not include error bars or formal significance tests.

\item {\bf Experiments compute resources}
    \item[] Question: For each experiment, does the paper provide sufficient information on the computer resources (type of compute workers, memory, time of execution) needed to reproduce the experiments?
    \item[] Answer: \answerYes{}
    \item[] Justification: We report the GPU type, memory budget, batch
    sizes, per-epoch time, estimated training time, and OOM or timeout cases.
    
\item {\bf Code of ethics}
    \item[] Question: Does the research conducted in the paper conform, in every respect, with the NeurIPS Code of Ethics \url{https://neurips.cc/public/EthicsGuidelines}?
    \item[] Answer: \answerYes{}
    \item[] Justification: We use public, aggregated OD records under the
    source data license. The released data do not contain private or
    personally identifiable information.

\item {\bf Broader impacts}
    \item[] Question: Does the paper discuss both potential positive societal impacts and negative societal impacts of the work performed?
    \item[] Answer: \answerNA{}
    \item[] Justification: The paper introduces a public, aggregated mobility
    dataset and OD forecasting benchmark. We do not identify a direct
    societal impact from the dataset and benchmark themselves.
    
\item {\bf Safeguards}
    \item[] Question: Does the paper describe safeguards that have been put in place for responsible release of data or models that have a high risk for misuse (e.g., pre-trained language models, image generators, or scraped datasets)?
    \item[] Answer: \answerNA{}
    \item[] Justification: The paper does not release high-risk generative
    models, scraped content, or individual-level mobility data.

\item {\bf Licenses for existing assets}
    \item[] Question: Are the creators or original owners of assets (e.g., code, data, models), used in the paper, properly credited and are the license and terms of use explicitly mentioned and properly respected?
    \item[] Answer: \answerYes{}
    \item[] Justification: We identify the Seoul Open Data source and KOGL
    Type 1 license, specify CC BY-NC 4.0 for the processed \dataset release,
    and credit external code components with source URLs and citations in the
    released repository.

\item {\bf New assets}
    \item[] Question: Are new assets introduced in the paper well documented and is the documentation provided alongside the assets?
    \item[] Answer: \answerYes{}
    \item[] Justification: The dataset and code releases include README files,
    metadata, schema descriptions, and scripts for data preparation and
    benchmark runs.

\item {\bf Crowdsourcing and research with human subjects}
    \item[] Question: For crowdsourcing experiments and research with human subjects, does the paper include the full text of instructions given to participants and screenshots, if applicable, as well as details about compensation (if any)? 
    \item[] Answer: \answerNA{}
    \item[] Justification: The paper does not involve crowdsourcing or
    direct human-subject experiments.

\item {\bf Institutional review board (IRB) approvals or equivalent for research with human subjects}
    \item[] Question: Does the paper describe potential risks incurred by study participants, whether such risks were disclosed to the subjects, and whether Institutional Review Board (IRB) approvals (or an equivalent approval/review based on the requirements of your country or institution) were obtained?
    \item[] Answer: \answerNA{}
    \item[] Justification: The paper does not involve human-subject
    experiments.

\item {\bf Declaration of LLM usage}
    \item[] Question: Does the paper describe the usage of LLMs if it is an important, original, or non-standard component of the core methods in this research? Note that if the LLM is used only for writing, editing, or formatting purposes and does \emph{not} impact the core methodology, scientific rigor, or originality of the research, declaration is not required.
    \item[] Answer: \answerNA{}
    \item[] Justification: LLMs were used only for editing and code
    refactoring, and were not used as an important component of the core
    methods in this research.

\end{enumerate}

\end{document}